\documentclass{article}

\usepackage{arxiv}

\usepackage[utf8]{inputenc} % allow utf-8 input
\usepackage[T1]{fontenc}    % use 8-bit T1 fonts
\usepackage{hyperref}       % hyperlinks
\usepackage{url}            % simple URL typesetting
\usepackage{booktabs}       % professional-quality tables
\usepackage{amsfonts}       % blackboard math symbols
\usepackage{nicefrac}       % compact symbols for 1/2, etc.
\usepackage{microtype}      % microtypography
\usepackage{lipsum}
\usepackage{amsmath}

\usepackage{subcaption}
\usepackage{graphicx}
\usepackage{hyperref}

\title{Binary autoencoder with random binary weights}

\author{
  Viacheslav M.~Osaulenko\\
  Igor Sikorsky Kyiv Polytechnic Institute\\
  Kyiv, Ukraine\\
  \texttt{osaulenko.v.m@gmail.com} 
  }

\renewcommand{\vec}[1]{\mathbf{#1}}
\newcommand{\matr}[1]{\mathbf{#1}}   

\newcommand{\euqall}[1]{\ensuremath{\stackrel{\text{#1}}{=}}}
\newcommand{\less}[1]{\ensuremath{\stackrel{\text{#1}}{ \leq}}}

\DeclareMathOperator{\argminG}{arg\,min}
\DeclareMathOperator{\argmaxG}{arg\,max}
\begin{document}
\maketitle

\begin{abstract}
Here is presented an analysis of an autoencoder with binary activations $\{0, 1\}$ and binary $\{0, 1\}$ random weights. 
Such set up puts this model at the intersection of different fields: neuroscience, information theory, sparse coding, and machine learning. 
It is shown that the sparse activation of the hidden layer arises naturally in order to preserve information between layers.
Furthermore, with a large enough hidden layer, it is possible to get zero reconstruction error for any input just by varying the thresholds of neurons.
The model preserves the similarity of inputs at the hidden layer that is maximal for the dense hidden layer activation. 
By analyzing the mutual information between layers it is shown that the difference between sparse and dense representations is related to a memory-computation trade-off.  
The model is similar to an olfactory perception system of a fruit fly,
and the presented theoretical results give useful insights toward understanding more complex neural networks.
\end{abstract}
% keywords can be removed

\section{Introduction}
"Rate vs spike coding" is an old debate \cite{Gerstner1997,Brette2015} over how information is encoded in biological neural networks (NN):
as a firing rate of action potentials or as individual spikes. The first approach is used in artificial NN where the output of a neuron is a real number. The second is used in spiking NN and association memory models where the output of a neuron is binary. 
A strong argument in favor of the second approach is that for human visual perception the neural activity passes through 5-7 areas with just 100ms \cite{VanRullen2001a}. Taking into account spike propagation (10-20ms) and generation (2-5ms) there is no time to transmit multiple spikes as a rate of firing. Instead, the stimulus most likely is encoded as a large sparsely active neural population \cite{Olshausen2004,Quiroga2008} that at first approximation can be represented as a binary vector. Thus, it is important to understand how such binary vectors are formed from arbitrary inputs. 

A simple way to encode information into a binary vector is to multiply an input vector by a random matrix (that is called a random projection) and apply threshold non-linearity. In the work \cite{Dasgupta2017} it is shown that such procedure is realized in the fruit fly brain to make representations of odors.
The signal from 50 odorant receptor neurons is transmitted to 50 projection neurons and further projected to a much larger layer with 2000 Kenyon cells (KC). The weights of synapses from projection neurons are random and binary $\{0, 1\}$.
The activity of KC is sparse, only around 5\% of neurons active at any given time.
The author assumed that KC sparsity is important since the signal further propagates to association areas and many works showed that the association memory capacity is maximal for sparse activation \cite{Knoblauch2010,Palm2013}.
Is a random projection a good way to represent the input? How much information is preserved? Why the encoding layer is sparsely active and much larger than the input?

This paper addresses these questions and treats the random projection with some non-linearity as an encoder of the autoencoder model. The added decoder tries to reconstruct the input from the hidden layer to achieve the lowest error. All activations and weights are binary $\{0,1\}$. 
Three different activation functions are used: a simple threshold model, a k-winners-take-all model, and a binary matching pursuit model. 
It is showed that for a large enough hidden layer it is possible to get zero reconstruction error for any input just by varying two parameters: the threshold of neurons in the hidden layer and the threshold in the output layer. 
With the increase of the hidden layer, the activation becomes more sparse in order to decrease the error.
All three models show the similarity preservation, but better reconstruction does not necessarily 
give better similarity preservation.
The best similarity preservation is achieved for 50\% active hidden neurons, much more than in the fruit fly.
Why is that? 

To understand why it is extremely useful to calculate mutual information. 
It is shown that the mutual information maximization and the error minimization is similar but not the same.
The mutual information between the input and the hidden layer is maximal for the dense hidden layer activation. 
But between the input and the output layers the mutual information is maximal for the sparse hidden layer activation. 
It is shown, that the reason to project the input to a larger dimension and to make it sparse is exactly to maximize the mutual information between the input and the higher neural layers.
Furthermore, the way an activation function is parametrized (like the linear weighted sum of inputs) plays a crucial role. Change to a bit complicated sigma-pi activation function that includes dendritic computation preserves more information. So, with more parameters and computation in a single neuron it is possible to make the overall network smaller and preserve the same information as with the large network but simple activation function (it is called here the memory-computation trade-off).

At the end, the paper makes a connection to (surprisingly) a Hopfield model by showing the presence of attractors. 
At the discussion section, some open problems are formulated that point to the promising directions of future research. 
 
The paper has an appendix, where an interested reader will find the mathematical analysis of the binary autoencoder, and understand how to calculate the average error analytically. Also, there is a mathematical derivation of mutual information and some interesting results that are put outside the main text to increase the overall clarity.

\section{The threshold model}
Let $\vec{x}, \vec{x^r} \in \{0,1\}^{N_x}$ -- binary vectors that represent input and output layers, $\vec{y} \in \{0,1\}^{N_y}$ -- binary hidden layer. 
Activation of the hidden (encoder) and the output (decoder) layer is given by:

\begin{align}
&\vec{y} = \theta(\matr{w} \vec{x} - t_y) \\
&\vec{x^r} = \theta(\matr{w}^T \vec{y} - t_x)
\end{align}

Where $\matr{w} \in \{0,1\}^{N_y \times N_x}$ -- weight matrix and $\theta(x) = \{1, \text{if } x \geq 0; 0,  x < 0 \} $ is a step function.  Here $t_y$ and $t_x$ are scalars, so the difference and the step function are applied element-wise. It means that all neurons in a layer share a single threshold. 
See the appendix \ref{model_variations} for why it is better to have the same threshold and dynamically change it for each input rather than to have fixed but individual thresholds.
The reconstruction error is given by the Hamming distance scaled by $N_x$: 

\begin{equation}
E(\vec{x}, \vec{x^r}) = \frac{1}{N_x} \sum_i{|x_i - x^r_i|} 
\end{equation}

Note that for binary vectors the Hamming distance is the same as $L_1$ distance and $L_0$ "norm".
For binary parameters the distance takes integer values thus it is not differentiable and traditional gradient descent optimization is not applicable.
Still, for binary variables the thresholds $t_y, t_x$ are integers and it is possible to iterate through all  threshold values to find the minimal error.  
The weights are initiated randomly such as each row has exactly $a_w$ ones, 
$\sum_j{w_{ij}} = a_w$ (see at the appendix \ref{model_variations} explanation why it is better than using Bernoulli distribution  $w_{ij} \in Ber(s_w) $ ).

This model does not consider a particular dataset, but treats the input as a random vector that 
has $a_x$ ones: $\sum_i{x_i} = a_x$. Also, analyzing different values of $t_x$ is less interesting 
compared to $t_y$, so for all experiments the optimal value $t_x = \argminG_{t_x}(E(\vec{x}, \vec{x^r}))$ is used.
Finally, the goal is to analyze how the error depends on five parameters $N_x, N_y, a_x, a_w, t_y$. 

\subsection*{Experiments}
All experiments are set up with parameters $N_x=50, N_y=150, a_x=20, a_w=30$ if other is not specified.

\begin{figure}[t!]
  \begin{subfigure}[t]{0.475\textwidth}
        \includegraphics[height=2.5in]{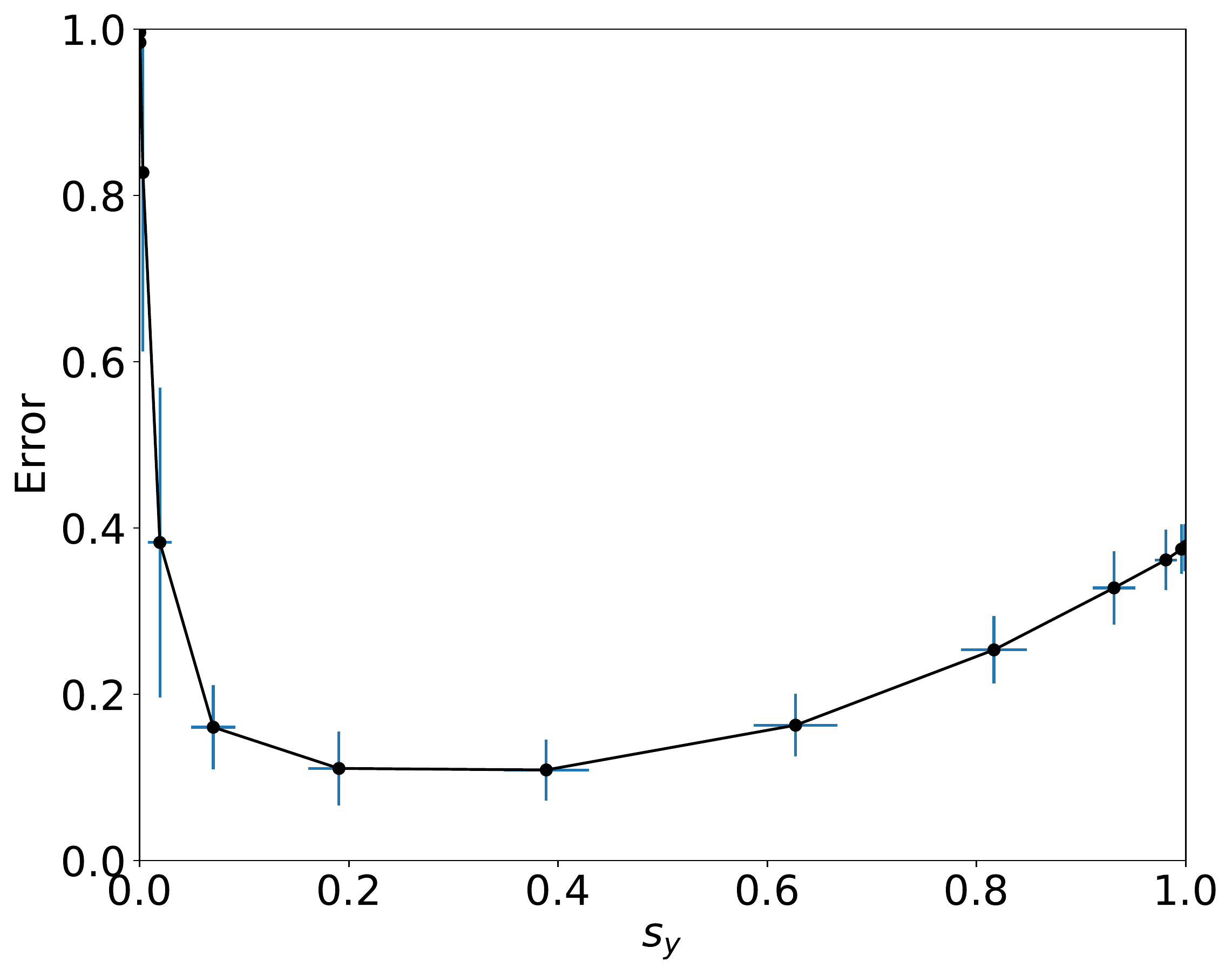}
      \caption{\label{fig:error_vs_ay_theta} }  
  \end{subfigure}
  \qquad
  \begin{subfigure}[t]{0.475\textwidth}
  
      \includegraphics[height=2.5in]{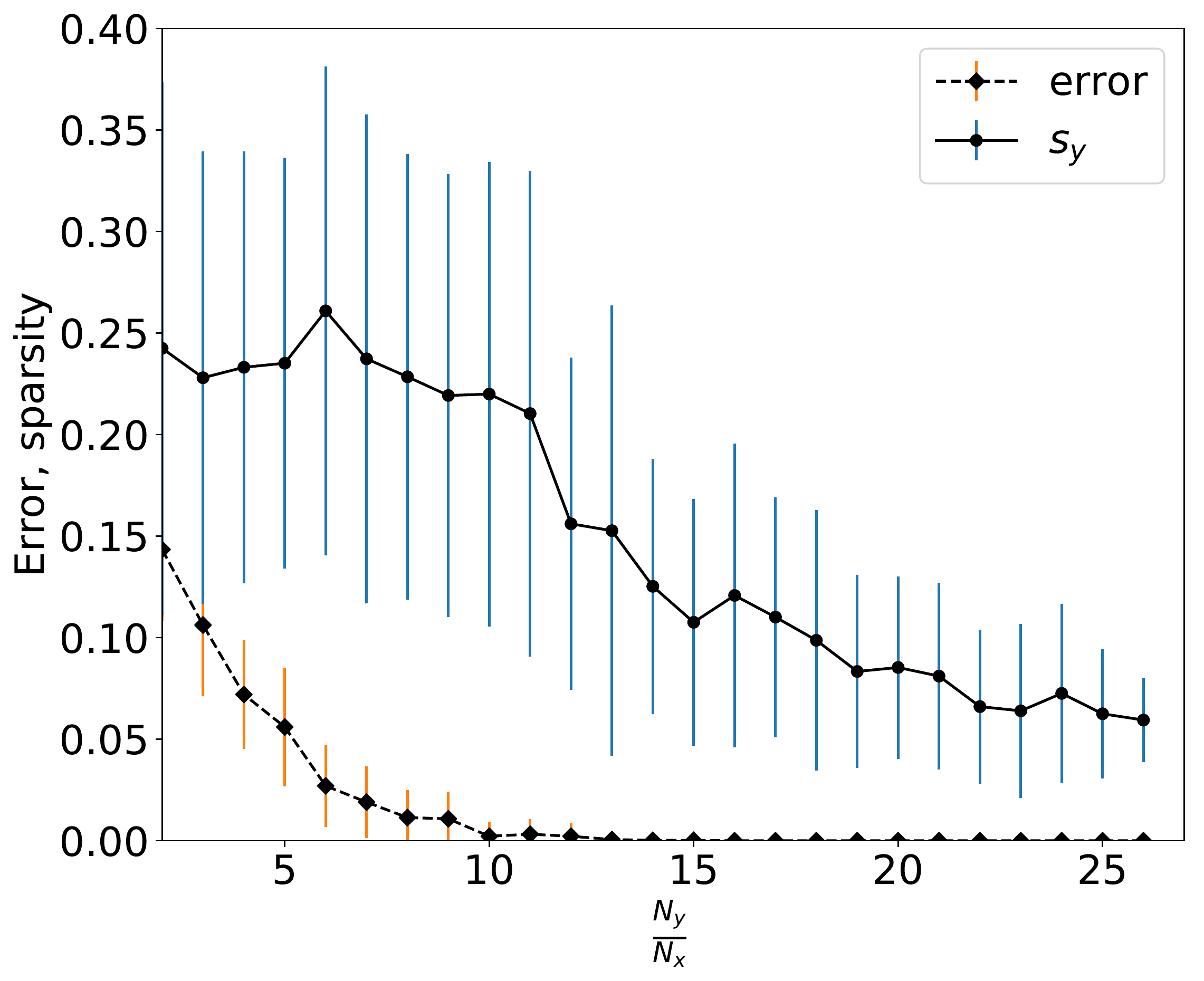}
      \caption{\label{fig:sy_vs_Ny_theta}}
  \end{subfigure}
  
    \begin{subfigure}[t]{0.475\textwidth}
  
      \includegraphics[height=2.5in]{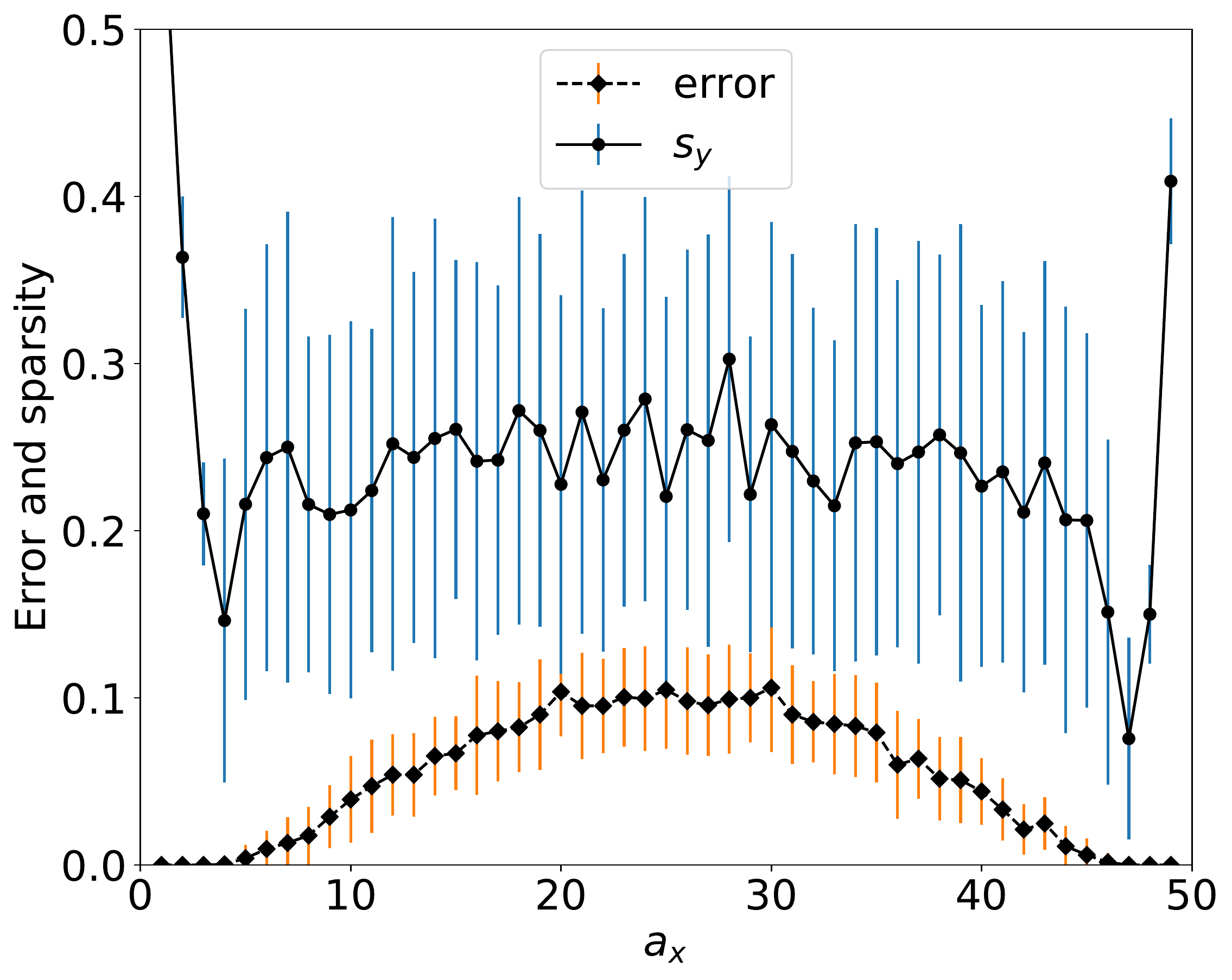}
      \caption{\label{fig:ay_vs_ax_theta}}
  \end{subfigure}
  \qquad
     \begin{subfigure}[t]{0.475\textwidth}
  
      \includegraphics[height=2.5in]{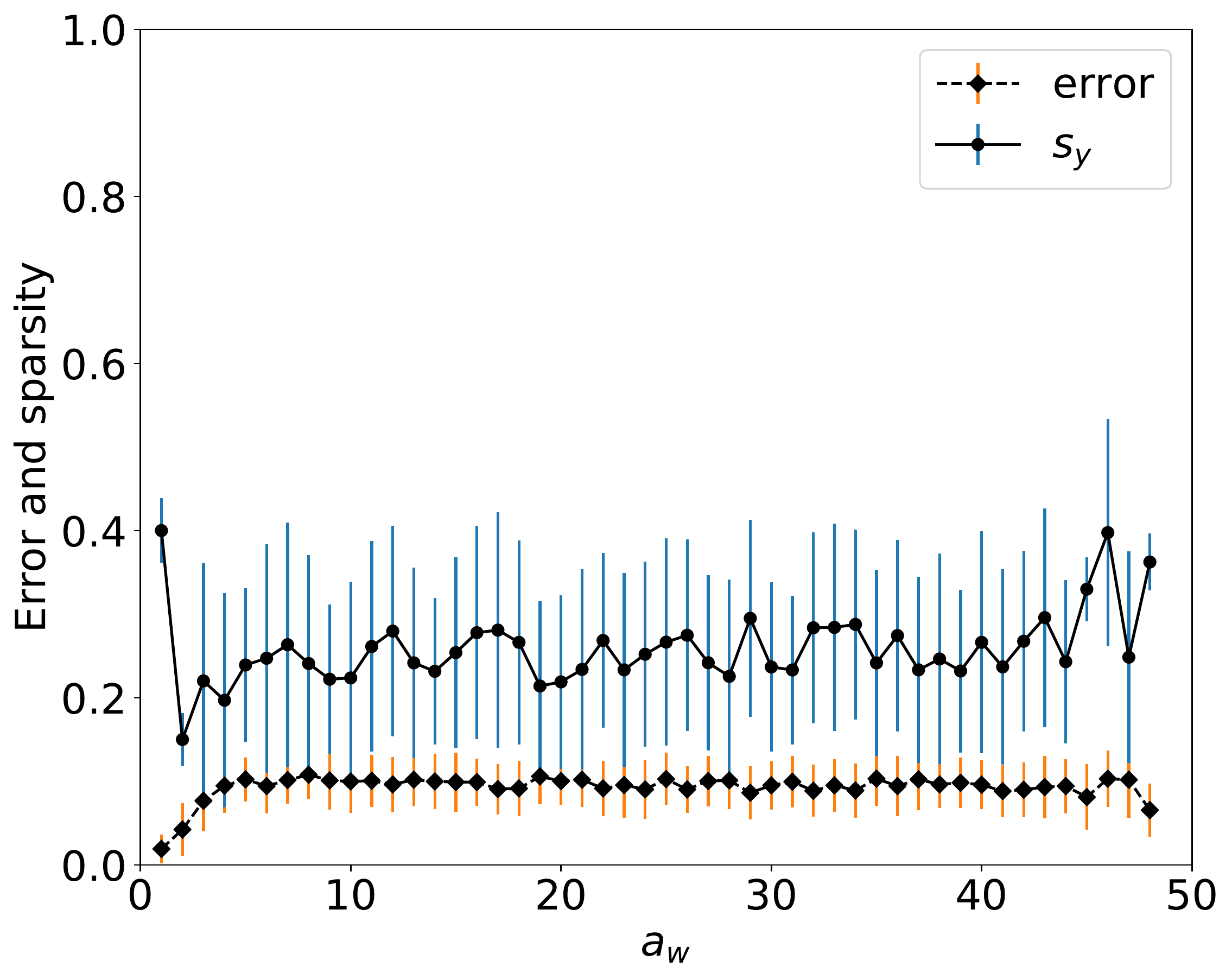}
      \caption{\label{fig:ay_vs_aw_theta}}
  \end{subfigure}
   
     \caption{Results averaged over random weights initializations for the threshold model. Standard deviation is specified with bars. 
          a) Reconstruction error vs hidden layer sparsity.   
          b) Reconstruction error and optimal hidden layer sparsity vs fraction $N_y/N_x$.
          c) Reconstruction error and optimal hidden layer sparsity vs the number of non zero elements in the input vector ($a_x$).
          d) Reconstruction error and optimal hidden layer sparsity vs the number of non zero elements in the row of the weight matrix  ($a_w$).
          	}
\end{figure}
The first experiment shows how the error depends on the threshold $t_y$. 
Each value of a threshold activates a certain number of hidden neurons $a_y$. 
It is convenient instead of $t_y$ to analyze the sparsity of a hidden layer $s_y = a_y/N_y$.
The fig.\ref{fig:error_vs_ay_theta} shows with circles the error for different values of thresholds $t_y$  (converted to corresponding sparsity on the x-axis).
The error has a minimum at around 20\% - 30\% of active hidden neurons that we will call an optimal sparsity. 
If the error is minimal (for example equals to zero) at many different sparsity levels, then the optimal sparsity is the smallest sparsity where the error is minimal. 
The experiment was repeated for different random weights with fixed $a_w$ and the averaged result presented on the plot. 
Note, it is the same as repeat the experiment for fixed weights but different random input with $a_x$ ones.
The bars show the standard deviation for an error and for the sparsity since for every random initialization the same threshold gives a different error and the number of active neurons.

Next experiment analyzes how the error depends on $N_y$. 
If the input $a_x/N_x$ and weight sparsity  $a_w/N_x$  are constant, then the error and optimal sparsity depend only on fraction $N_y/N_x$. 
Fig.\ref{fig:sy_vs_Ny_theta} shows that the minimum error (dash line) and the optimal sparsity (solid line) decrease while the size of a hidden layer increases.  
With the large enough hidden layer, it is possible to achieve zero reconstruction error with random weights just by changing thresholds for any input vector.

The solid line on fig.\ref{fig:ay_vs_ax_theta} shows that the error decreases
when the number of active input neuron $a_x$ approaches $0$ or $N_x$
and is maximal at $a_x = N_x/2$. 
The dashed line shows that optimal sparsity remains relatively in the 20\% - 30\% window, but for sparse inputs $a_x <5, a_x > 45$ the optimal sparsity increases.
 
Finally, the fig.\ref{fig:ay_vs_aw_theta} shows with the solid line that the error and optimal sparsity remains relatively constant for a different number of positive weights $a_w$ (except for boundary values).

\section{k-winners-take-all and binary matching pursuit models}
The threshold in the previous model takes integer values, and increase by one decreases the number of active neurons by much more than one (fig. \ref{fig:error_vs_ay_theta}).  
To calculate the error for intermediate sparsity levels we can modify the model by replacing the threshold non-linearity with k-winners-take-all:

\begin{align}
&\vec{y} = kWTA(\matr{w} \vec{x}, a_y) \\
&\vec{x^r} = kWTA(\matr{w}^T \vec{y}, a_x^r)
\end{align}  

The function $kWTA(\vec{z}, k)$ takes vector $\vec{z}$ and returns $k$ highest values as ones and others as zeros.
For example, $kWTA([1,4,3,2,5], 2) =[0,1,0,0,1]$.
Thus, we can estimate the error at more points $a_y \in [1, N_y]$ compared to $t_y \in [1, \min(a_x, a_w)]$.
As in the previous model, the parameter $a_x^r$ is selected to minimize the error.
However, it is worth noting that on average it equals the number of ones in the input $a_x^r = a_x$  (see \textit{average\_a\_x.py} at \cite{Github2020}). 
So, to speed up the reconstruction, we can skip the iteration of all
 possible $a_x^r$, but that results in a slightly larger error.

Usually, the element of the activation vector before non-linearity $\vec{z} = \matr{w} \vec{x}$
is called an overlap.
For binary vectors the overlap is an integer and it may be not possible 
to select exactly $a_y$ most active neurons. For example, $kWTA([2,2,3], 2)$ can give $[0,1,1]$ or $[1,0,1]$.
We can set ambiguous indices to zero at random or by the first occurrence. 
For simplicity the later is used, since the two approaches give the same results if the error is averaged over random weights.
Note, if the weights are real numbers, then the overlap is a real number, so the $kWTA(\matr{w} \vec{x},a_y)$ is identical to the threshold model $\theta(\matr{w} \vec{x} - t_y)$ where $t_y$ is selected to output exactly $a_y$ non-zero elements.

As presented next, the results for kWTA model give a lower error and more sparse activations compared to the threshold model. 
Still, we can step aside from traditional weighted inputs $\matr{w} \vec{x}$  to a more algorithmic approach inspired by greedy algorithms from the sparse coding field \cite{Elad2010,Elad2015}. 
At a first step, we can select the best weight vector $\vec{w_i}$ that reconstructs the input $\vec{x^r_1}$  (it has the highest overlap). 
At a second step, select the new weight vector that has the highest overlap with the residual $2\vec{x} - \vec{x^r_1}$. Vector difference prevents from selecting the weights that have high overlap with the false positive ($x_i=0, x^r_i=1$) elements and the multiplication by 2 reduces the false negatives ($x_i=1, x^r_i=0$).
Repeat the procedure for $n$ step  and get the reconstruction vector  $\vec{x^r_n}$.
In this paper this algorithm is called a binary matching pursuit and can be expressed as:

\begin{align}
& \vec{y_n} = \vec{y_{n-1}} + kWTA(\matr{w} ( 2\vec{x} - \vec{x^r_{n-1}}) - \lambda \vec{y_{n-1}}, 1) \\
&\vec{x^r_n} = kWTA(\matr{w}^T \vec{y_n},  a_x^r)
\end{align}

where $\vec{y_0} = \vec{0},\vec{x^r_0}=\vec{0}$ - zero vectors, 
$n$ - the step number and the number of active neurons in hidden layer,
the residual $2\vec{x} - \vec{x^r_{n-1}}$ defines the next weight vector,
the parameter $\lambda > 2a_x a_w$ ensures that at each step we do not select already selected weight vector,
 $kWTA(\vec{x}, 1)$ returns a one-hot vector where the one corresponds to the largest input value. 

\subsection*{Experiments}
The fig.2 shows results for the same experimental setup and same parameters as in fig.1 but with three models together.
The results are divided into two set of figures to show the standard deviation for the first model (otherwise the figure would be clattered). Also, it is better
to familiarize the reader with the main results on a simple model subsequently showing that there are better choices.  

\begin{figure}[t!]
  \begin{subfigure}[t]{0.475\textwidth}
        \includegraphics[height=2.5in]{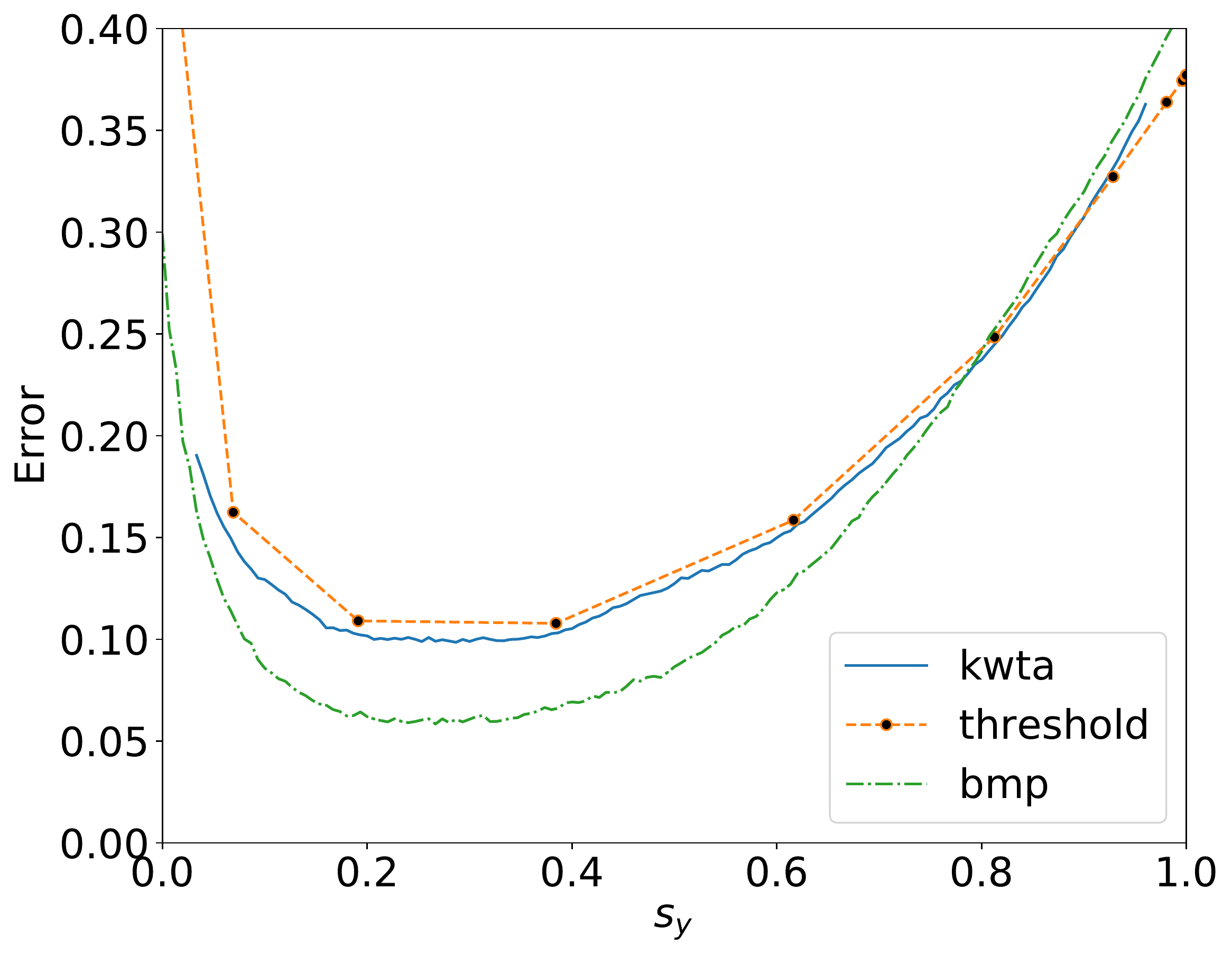}
      \caption{\label{fig:error_vs_ay_all} }  
  \end{subfigure}
  \qquad
  \begin{subfigure}[t]{0.475\textwidth}
        \includegraphics[height=2.5in]{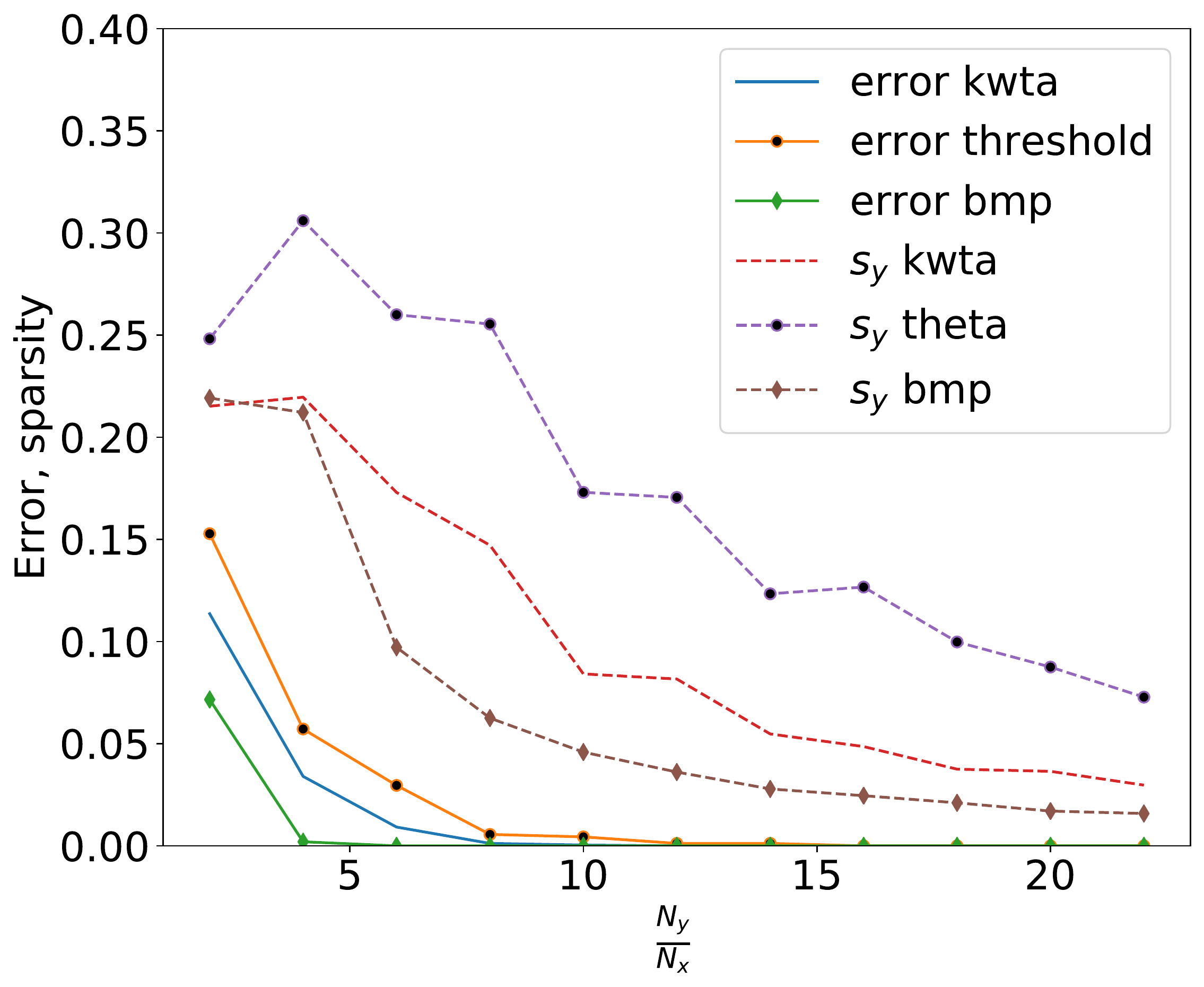}
      \caption{\label{fig:error_sy_vs_Ny_all}}
  \end{subfigure}
  \qquad
     \begin{subfigure}[t]{0.475\textwidth}
       \includegraphics[height=2.5in]{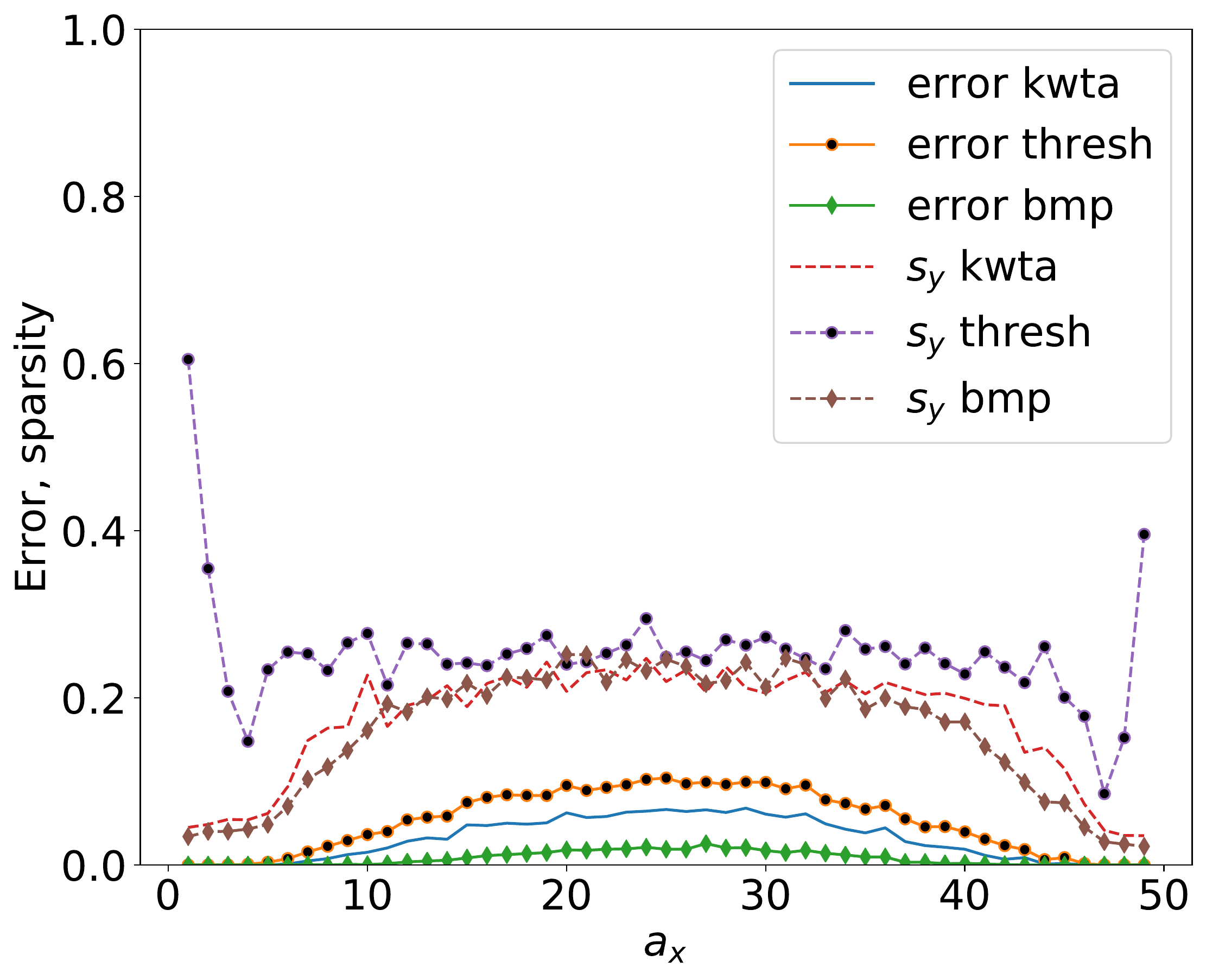}
      \caption{\label{fig:error_sy_vs_ax_all}}
  \end{subfigure}
       \begin{subfigure}[t]{0.475\textwidth}
        \includegraphics[height=2.5in]{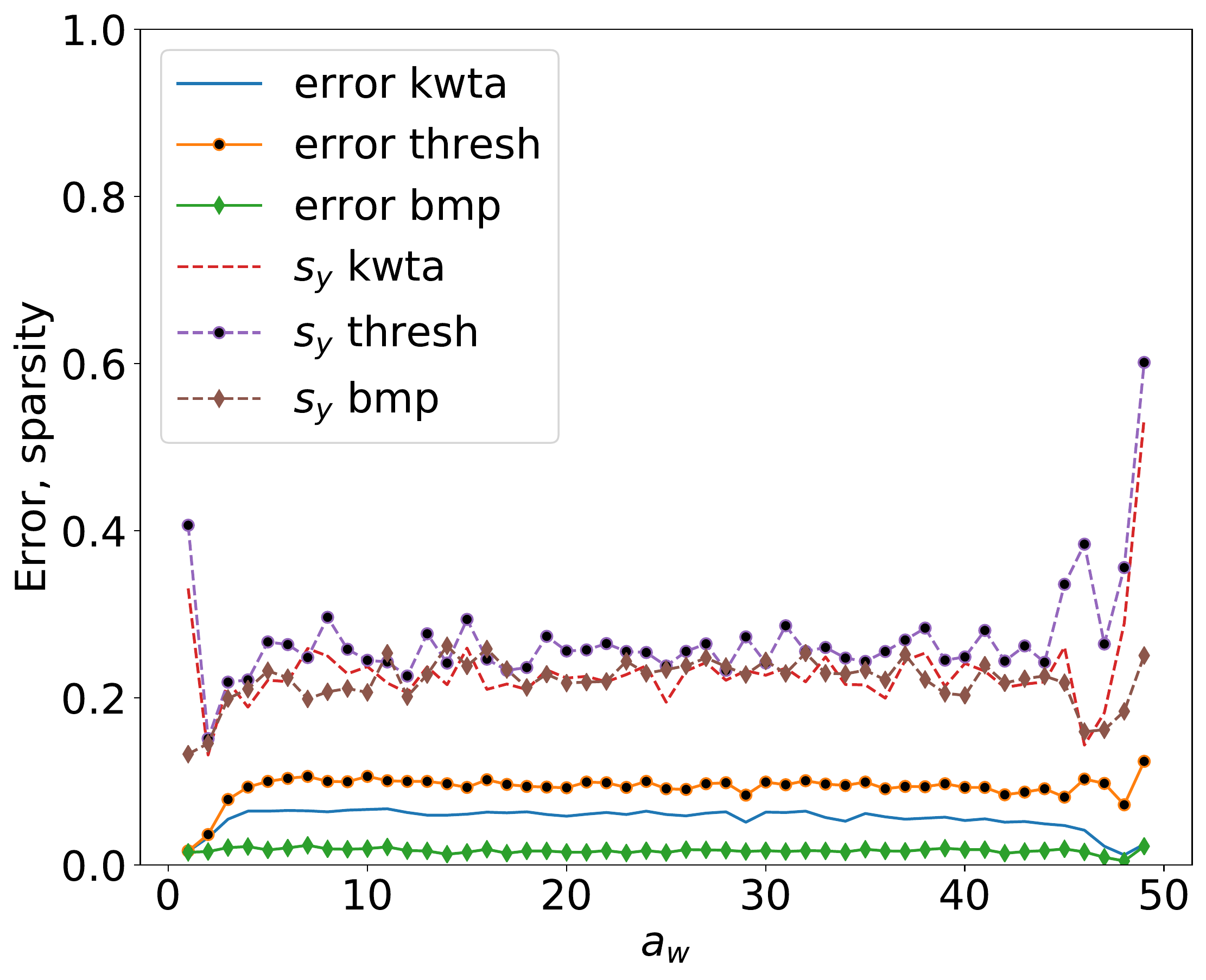}
      \caption{\label{fig:error_sy_vs_aw_all}}
  \end{subfigure}   
     \caption{
           Results averaged over random weights initializations for the threshold, kWTA and BMP models. 
          The axes for all graphs are the same as for the fig.1. See the legend to relate the curve to the model.
          	}
\end{figure}

On the fig. \ref{fig:error_vs_ay_all} we see that the kWTA model fills intermediate points between thresholds, though it is still discrete. The average errors for the two first models are very close, showing their similarity. Binary matching pursuit model (BMP) has better results compared to the two. 

Next, fig. \ref{fig:error_sy_vs_Ny_all} presents how the average minimum error and the optimal sparsity depend on $N_y/N_x$ for three models.
Note, that the previous figure showed the average error for fixed $s_y$, but here the minimal error is averaged over different sparsity values. 
That is why here the error for the kWTA model is different from the threshold model (for the previous figure they were almost the same). 
It is because the kWTA calculates the error in a larger interval $a_y \in [1, N_y]$ versus $t_y \in [1, \min(a_x, a_w)]$ 
and it is more likely to get the lower minimal error at each random initialization.
The BMP model has the lowest error and approximately for $N_y > 3N_x$ the error is zero 
and the sparsity is lower than 5\%. 
Not shown here, but the number of active hidden neurons $a_y$ also decreases with the increase of the hidden layer. 
For the kWTA model approximately with $N_y/N_x = 10$ we have $a_y \approx N_x$ and for $N_y/N_x > 50$  we have $a_y < a_x$ 
 (see the code \textit{high\_ratio.py} \cite{Github2020}).
Consider the extreme case of local coding when $N_y = 2^{N_x}$ 
and weights are equal to all possible input vectors.
Then for any input, the reconstruction error is zero and the hidden layer has just one active neuron. 
Read more at appendix (\ref{intuition_reconstruction}).

Similarly, on fig. \ref{fig:error_sy_vs_ax_all} and fig. \ref{fig:error_sy_vs_aw_all} BMP model shows the lowest error and sparsity, though the difference in sparsity is not so big.
Also, note that for sparse inputs the kWTA and BMP models give more sparse encodings.

\section{On the similarity preservation}
kWTA activation function is inspired from biology, where lateral inhibition selects winning excitatory population that encodes the input. 
Such procedure is present in the fruit fly for KC layer, mentioned in the introduction, 
where single inhibitory neuron controls the sparsity level.
The kWTA function with random weights preserves similarity: similar smells are encoded as similar neural activation patterns. 
It provides a simple form of a generalization: if a new smell is similar to some that previously led to a positive reinforcement, 
the fly will pursue the source of the smell.

Similarity preservation for binary  $\{0, 1\}$  hidden layer and weights and real-valued input was studied before \cite{Rachkovskij2015,Rachkovskij2015a}.
But the questions of hidden layer sparsity and error reconstruction were not addressed. 

In the fruit fly, the sparsity of the hidden layer is about 5\%, and layer sizes $N_x=50, N_y=2000$ with the ratio equal to 40. 
In the fig. \ref{fig:error_sy_vs_Ny_all} we see that this level of sparsity approximately gives the lowest error of reconstruction
(as if a fruit fly optimizes the reconstruction).
But how the hidden layer sparsity influences the similarity preservation? 

\begin{figure}[t!]
  \begin{subfigure}[t]{0.475\textwidth}
        \includegraphics[height=2.5in]{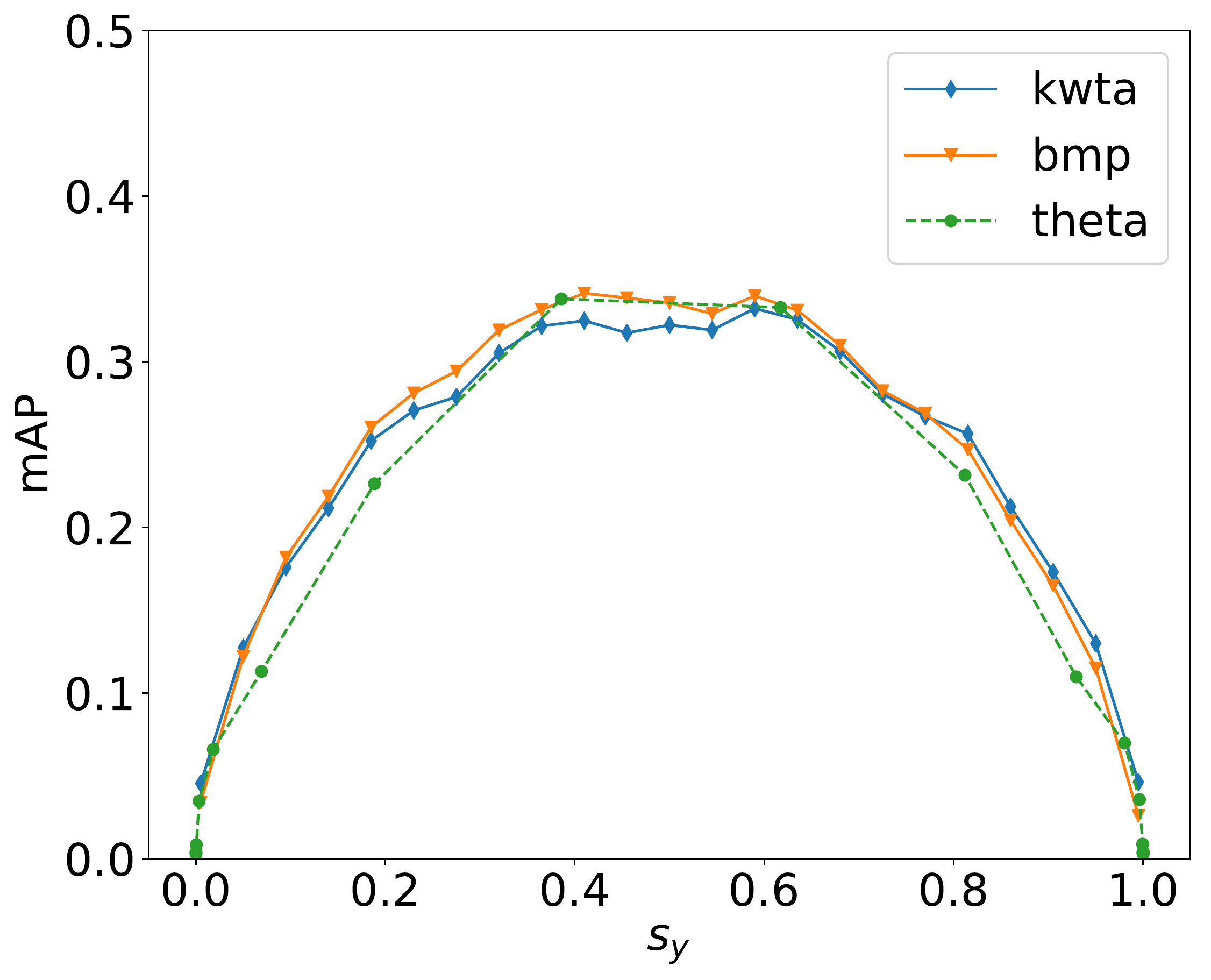}
      \caption{\label{fig:similarity} }  
  \end{subfigure}
  \qquad
  \begin{subfigure}[t]{0.475\textwidth}
  
      \includegraphics[height=2.5in]{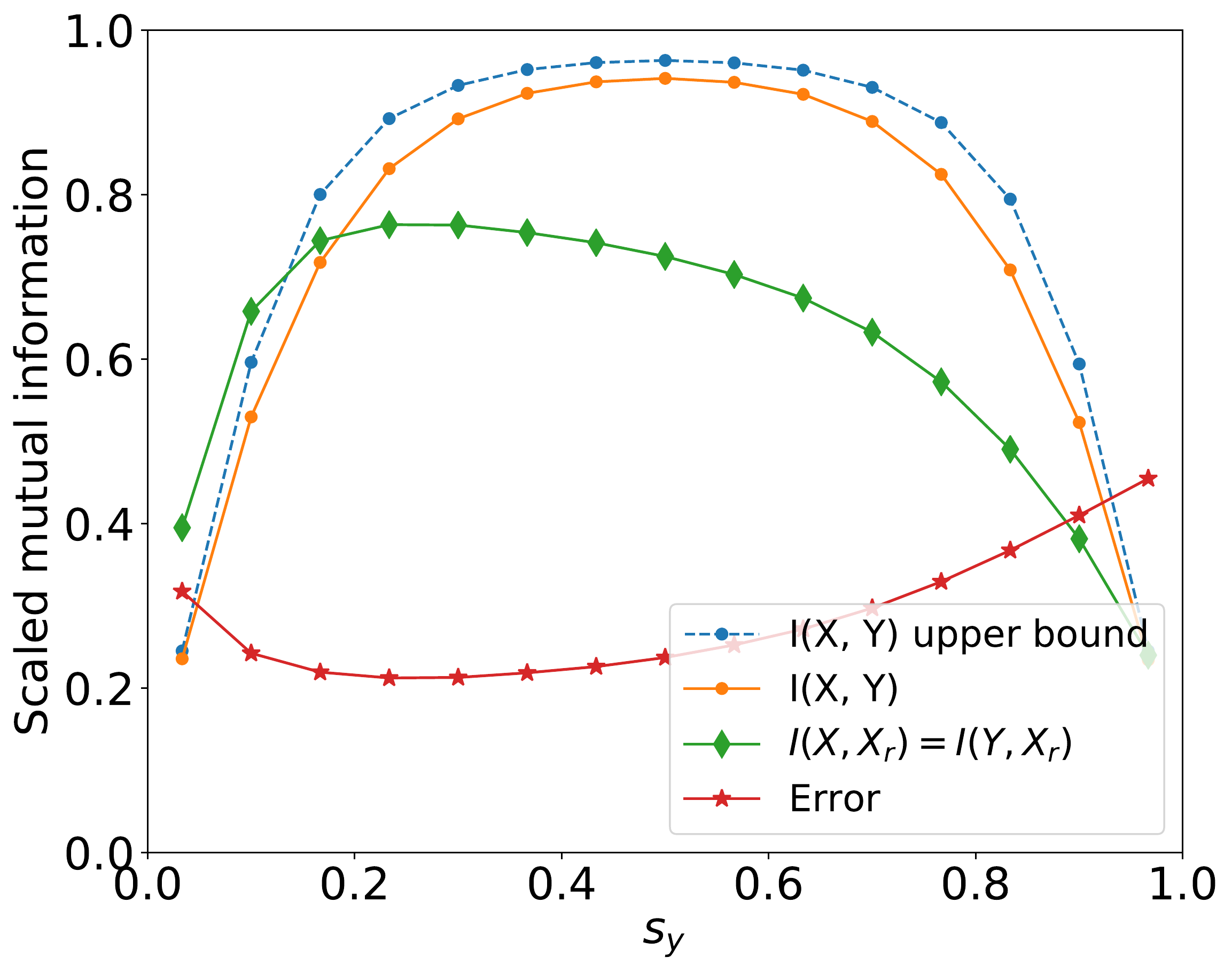}
      \caption{\label{fig:mutual_info}}
  \end{subfigure}  
   \qquad
     \begin{subfigure}[t]{0.475\textwidth}
       \includegraphics[height=2.5in]{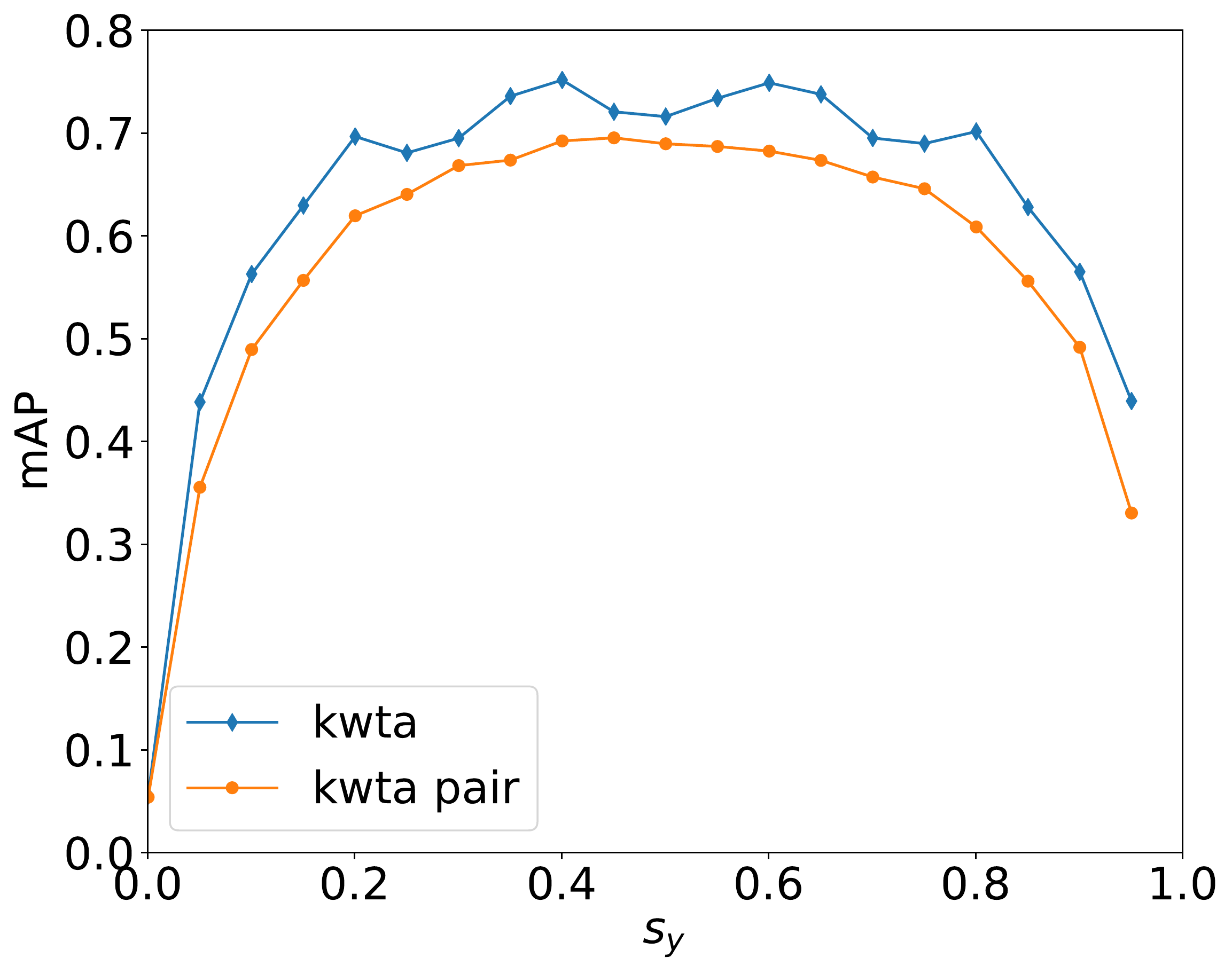}
      \caption{\label{fig:similarity_pair}}
  \end{subfigure}
       \begin{subfigure}[t]{0.475\textwidth}
        \includegraphics[height=2.5in]{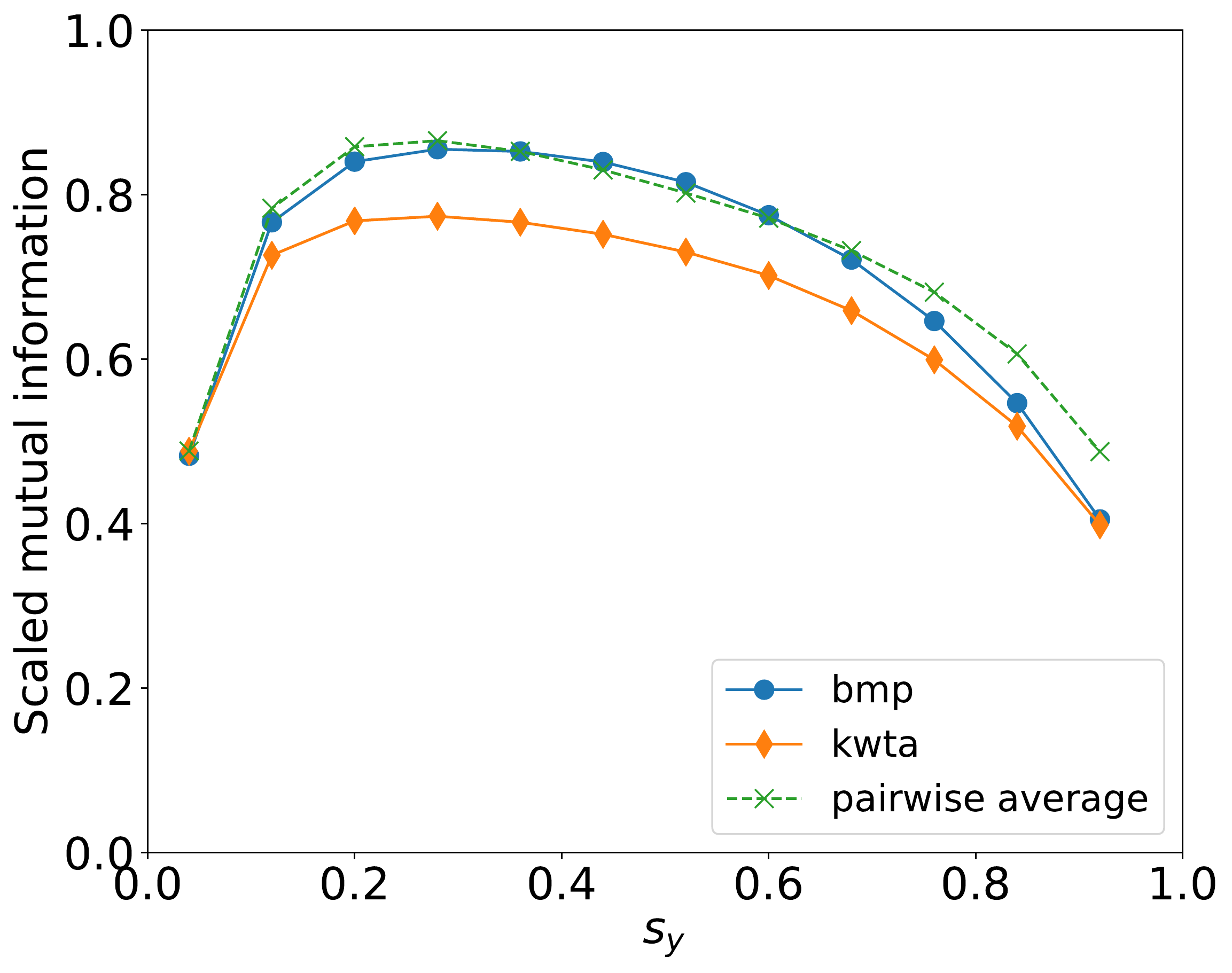}
      \caption{\label{fig:mutual_information_pair}}
  \end{subfigure}     
     \caption{
          a) The values of mean Average Precision (mAP) for different levels of the hidden layer sparsity. Results for three models are shown.
          b) Results of scaled MI for the kWTA model depending on the hidden layer sparsity. 
          Dashed curve is a scaled upper bound of MI between the input and the hidden layers ($I(X, Y)/H(X)$); 
          below is the scaled MI between the input and the hidden layers; next, below is the scaled MI between the hidden and the output layers $I(Y, X_r)/H(X)$; 
          the lower curve is the scaled average error of reconstruction.
          c) Mean Average Precision (mAP) for  the kWTA model (upper) and the kWTA model with pairwise correlations (lower). mAP is higher compared to the previous figure 
          due to the larger hidden layer size ($N_y = 2000$). 
          d)The scaled MI between the hidden and the output layers for the kWTA model (lower curve), the BMP model (blue, circles) and the kWTA with pairwise correlations (dashed line).
          	}
\end{figure}

The common measure of similarity preservation is a mean averaged precision (mAP). 
Roughly speaking, it shows to what extent similar vectors to some input vector remain similar in the hidden layer after encoding. 
The precise procedure of calculating the mAP for presented models is given at the appendix \ref{similarity_map}. 
The fig. \ref{fig:similarity} shows the dependence of mAP for three models for different levels of hidden layer sparsity. 
The maximal similarity preservation is achieved approximately at $s_y=0.5$.
On average, the BMP model gives better preservation than the kWTA model and kWTA better than the threshold model. 
At maximal values, the threshold model surprisingly gives comparable results to BMP model. 
So it can not be firmly stated that the better the model reconstructs the input, the better is the similarity preservation.
The graph raises the question, why the similarity preservation is the best at $s_y =0.5$, but the  reconstruction is the best at around $s_y=0.2$ (for $N_x=50, N_y=200$)? Note, the hidden/input layer size ratio for this graph is 4, far from 40 as in the fruit fly. 
   
Another important question is how the mAP is changing with $N_y$.
For random data when the $N_y > 50 N_x$ the mAP saturates approximately at $0.8$ level and never reaches the perfect similarity preservation (see \textit{similarity\_large\_N.py}).
The mAP is about 0.4 for the fruit fly parameters $s_y=0.05, N_y = 40 N_x $. 
It is higher than as shown on the fig.\ref{fig:similarity}, but still twice as low as optimal for such layer size. 
For completeness, the mAP does not significantly depends on $a_x$ and $a_w$ except for boundary values.
So we cannot tell that the sparsest the input the better the similarity preservation.

Also, it needs to mention, that the inputs for the fruit fly are not random, and it is questionable if the binary activation is a good approximation of sensory neuron activation. 
However, the question is still open, why activation of the hidden layer in the fruit fly promotes sparseness that gives worse similarity preservation? 
Maybe, it is more important to have high sparsity for better association memory, 
since the similarity preservation is worthless if it does not lead to correct behavior?

To understand the discrepancy between the reconstruction and similarity preservation let us try to analyze another measure -- the mutual information.

\section{On the mutual information}
The plan is the following: get the mutual information between the input and hidden layers and see at what sparsity levels it is maximal. 
Is it at 50\% or 5\% or at completely different value?
Intuitively, if the model can successfully reconstruct the input it preserves all information.

The mutual information (MI) between two random variables $X, Y$ can be calculated 
as a difference between the entropy of one variable and conditional entropy of this variable given the other:
$I(X,Y)=H(X)-H(X|Y)=H(Y)-H(Y|X)$.
We can treat the input vector as a random variable $X$ with uniform distribution ( $p_X(\vec{x}) = 1/2^{N_x}$) that has the entropy $H(X)=N_x$.
The hidden layer is a deterministic function of the input, thus it is also a random variable $Y$ with $p_{Y|X}(\vec{y}|\vec{x})=1$ (and $H(Y|X) = 0$ that just means that knowing the input we know the hidden state). 
Despite the function is deterministic not all information is preserved. 
It is because the activation function, for example the threshold model, can encode several input vectors into one hidden, 
thus $p_{X|Y}(\vec{x}|\vec{y}) \leq 1$ and  $H(X|Y) \geq 0$.
Also, note that the encoding does not realize all possible hidden states ($2^{N_y}$) (consider the case $N_x < N_y$), 
so the entropy of $Y$ can not be bigger than $H(X)$.
This functional mapping is called nonsurjective and noninjective.
In this case, the MI between input and hidden layers for uniform input distribution is given by:

\begin{equation}
\label{eq:mut_info}
I(x,y) = H(Y) = H(X) - \frac{1}{2^{N_x}}\sum_i^Z{|\Omega_i| log_2(|\Omega_i|)} \leq H(X) - log_2(\Omega^*)
\end{equation}

\begin{figure}[t!]
  \begin{subfigure}[t]{0.475\textwidth}
        \includegraphics[height=2.3in]{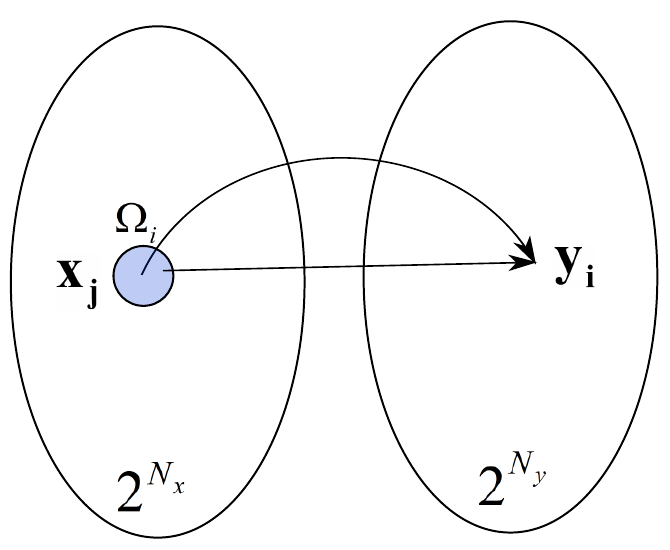}
      \caption{\label{fig:encoder} }  
  \end{subfigure}
  \qquad
  \begin{subfigure}[t]{0.475\textwidth}
  
      \includegraphics[height=2.3in]{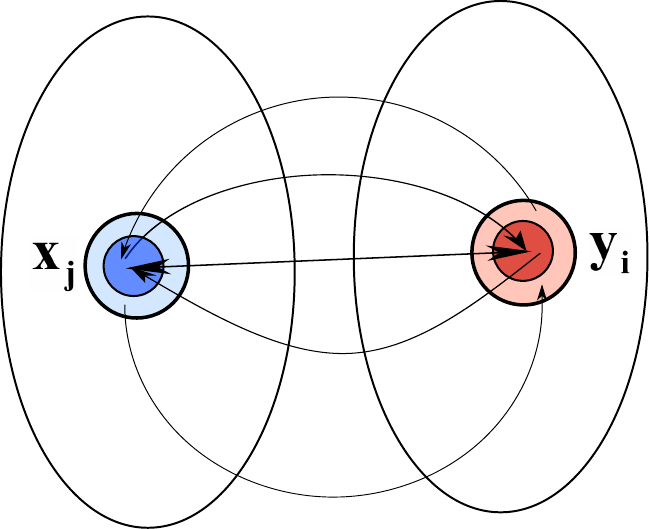}
      \caption{\label{fig:attractor}}
  \end{subfigure}    
     \caption{
          a) Projection from the input to the hidden space. All possible vector states are depicted.
           Blue area $\Omega_i$ shows all input vectors that are projected to the same hidden vector $\vec{y_i}$.
          b) Illustration of an attractor and the part of its basin. Arrows show the direction of the projection. 
          Darker colors show the set of inputs vectors, that lead to a single vector. 
          Lighter colors show the set of vectors that lead to another set in the opposite layer.  
                    	}
\end{figure}

Where $Z$ is the number of realized $\vec{y}$ vectors among total $2^{N_y}$,
 $\Omega_i = \{\vec{x_j}| \vec{y_i} = f(\vec{x_j}) \}$ -- a set of input vectors that lead to one hidden vector $\vec{y_i}$ ,
  $|\Omega_i|$ - number of elements in the set
 and $\Omega^* = E(|\Omega_i|)$ the mean number of elements in the set. 
 See the derivation at appendix \ref{mutual_info}. 
For better understanding see the fig. \ref{fig:encoder}.
Two sets represent all possible input hidden states ($2^{N_x}$ and $2^{N_y}$ respectively).
The light blue area depicts the $\Omega_i$, all input vectors that projects to one hidden. 

The equality in (\ref{eq:mut_info}) holds if the hidden state distribution is uniform, meaning $p_Y(\vec{y}) = 1/Z$ (for independent random variable the uniform distribution would be $p_Y(\vec{y}) = 1/2^{N_y}$ ).
The term with $\Omega_i$ corresponds to conditional entropy  $H(X|Y)$. 
If $\Omega^* \longrightarrow 1$ the mapping tends to become injective (one hidden vector corresponds to only one input vector) 
and the MI is maximal. 
 Also, in this case, the variance should tend to zero 
$Var(|\Omega_i|) \longrightarrow 0$.

One of the most important figures of the paper, the fig. \ref{fig:mutual_info}, shows how MI scaled by $H(X)$ depends on the hidden layer sparsity. 
The layers sizes $N_x=20, N_y = 30$, random weights with $a_w=7$, kWTA model. 
With just 20 input neurons it is possible to computationally enumerate all $2^{20}$ input vectors and calculate $\Omega_i$ for each vector $\vec{y}$. 
The top dashed line shows the upper bound (the $log_2\Omega^*$ expression) of the scaled MI between the input and the hidden layers. The orange line with circles shows the scaled MI between the input and the hidden layers.
Two lines are quite close, which means that random weights and uniform random inputs give close to uniform hidden vector distribution. 

The MI of an encoder ($X \rightarrow Y$) is the best at 50\% sparsity level, just like similarity preservation. 
But why the error is minimal at much lower sparsity?
The plot (fig. \ref{fig:mutual_info}) also shows with the green line the scaled MI between the input and the output layers that is the same as MI of the decoder (between the hidden and the output layers). 
The MI is the same because for the deterministic activation $X \rightarrow Y \rightarrow X_r$  we have $I(X,X_r)=I(Y,X_r)=H(X_r)$ (note that data processing inequality holds: $I(X,Y) \geq I(X, X_r)$).
The last red line shows the scaled error of reconstruction. 
So not only the error but the MI of the decoder is maximal at lower than 50\% of active hidden layer cells. 

Two lines reach the optimal value at the same sparsity,
but the MI maximization and the error minimization are not the same.
The first increases the number of states of the output layer and makes them equiprobable, 
while the second makes output states closer to input states. 
A good demonstrative example is if we take random decoder weights (instead of transpose).
In this case, the MI stays the same since the number of states is the same (on average). But the error rises because output states are different from inputs (see the code fig3b.py \cite{Github2020}). 
However, they both are optimal at the same sparsity because 
a high number of states is necessary to get a low error.

Finally, why the MI of a decoder is maximal for a sparse activation of a hidden layer?
The MI between two layers also depends on the input layer sparsity:  it is minimal at $a_x = N_x /2$ and maximal at $a_x=1$ or $a_x=N_x-1$. 
For the output layer the hidden layer act as an input, so one might expect that at $a_y=1$ or $a_y=N_y-1$ the MI of a decoder should be maximal. 
However, for $a_y=1$ all $2^{N_x}$ input vectors are compressed into $N_y$ 
hidden states with huge information loss, but each hidden state is decoded into unique output states. 
For $a_y=N_y/2$ the hidden layer looses minimal information but the output layer cannot realize all $2^{N_x}$ states.
Thus, there is some balance in the middle when the hidden layer should lose some information to prevent even more loss at higher layers.

The MI is determined by a particular function that connects two layers. For our task, there are $2^{N_y2^{N_x}}$ possible functions to map one vector space to another. 
The linear weighted sum with some nonlinearity $f(\matr{w}\vec{x})$ is just a one way of parametrization. It does not realize all possible functions and it is far from being the best (BMP gives better results).
For a quick example, we can use another biologically inspired activation function.
The typical pyramidal neuron has a large dendritic tree that can nonlinearly integrate the incoming activation due to dendritic spikes \cite{Sjostrom2008a,Branco2010}. 
A simplified model for such behavior is a sigma-pi neuron \cite{Mel1992}
 $y_j = f(\sum_k{w_k\prod_{i \in \Delta_k}{x_i}})$ where $\Delta_k$ - is the set of indices on a local 
dendritic branch. 
If we change the decoder from linear ($w_{ji}x_j$) to  to pairwise ($w_{ijk}x_jx_k$) summation inside kWTA nonlinearity, the MI increases (comparable to the BMP model, see the fig. \ref{fig:mutual_information_pair}).

With more parameters ($N_y \times N_x^2$) less information is lost.
This leads to the assumption that there is a trade-off between model size(number of neurons) and model complexity(number of parameters or required computation). 
Either you chose the complex function with more parameters and a small hidden layer, 
or you take simple weighted summation and increase the model size ($N_y$) to get the same MI.
Such trade-off is present in data compression and can be observed in biological neural networks.
Returning back to the fly brain, the hidden to input ratio is 40, however, 
the 5 (approximately) gives already maximal MI for the encoder and dense ($s_y=0.5$) hidden layer activation.
However, the higher areas use simple neural models and cannot preserve information well for the dense activation.
Thus, the hidden layer is made larger to preserve more information at higher layers with simple neural models.
It is an interesting direction of research to connect mutual information and neural activation function complexity measure (like the VC dimension \cite{Vapnik1994} ).

Lastly, a better encoder with higher MI does not imply better similarity preservation.
See the fig.\ref{fig:similarity_pair} where the pairwise correlation (lower curve) gives worse performance compared to linear kWTA
(graph for the $N_x =50, N_y=2000$ a fruit fly parameters). 
It is another important question to be answered elsewhere: how to choose a function that preserves both information and distance relationships?

\textbf{On attractors }
 
Consider another situation. Let us project some 
input vector $\vec{x_1}$ to $\vec{y_1}$ and back to $\vec{x_2}$. 
The error $E(\vec{x_1}, \vec{x_2}) \neq 0$ is not zero.
Imagine, that we repeat the procedure, project $\vec{x_2}$ to $\vec{y_2}$ and back to $\vec{x_3}$ and it appears that $\vec{x_3} = \vec{x_1}$.  
So, technically we received the best reconstruction but in two steps. 
Such trajectory $\vec{x_1}\rightarrow \vec{y_1} \rightarrow \vec{x_2} \rightarrow \vec{y_2} \rightarrow \vec{x_1}$ forms a limit cycle attractor of length 4, so that any other input or hidden vector that are projected to any of the trajectory elements will converge to the trajectory and stay there. 
Should we redo all experiments with the presented models with multi-step retrieval?
Preliminary computational experiments show that for presented models
there are no limit cycles other than length 2 ( though it requires theoretical proof).
Fig.\ref{fig:attractor} shows the attractor $\vec{x_j}\rightarrow \vec{y_i} \rightarrow \vec{x_j}$ and the part of the basin of attraction. 
Suppose some input vector belongs to the light blue area, it projects to light red, projects back to darkblue and darkblue projects to $\vec{y_i}$ that form the cycle with $\vec{x_j}$. The basin shows all input and hidden vectors that converge to the attractor.

The network architecture (layer sizes, weights, activation functions) defines the total number of vector pairs ($\vec{y_i}\leftrightarrow \vec{x_j}$) that form the attractors. 
It means that initially there are some set of input vectors that gives zero error reconstruction.
Other vectors have nonzero error and increase the average error value. 

Note that if for the threshold model we put $N_x = N_y$, use symmetric weights $\matr{w} = \matr{w^T}$, set  diagonal to zero $w_{ii} = 0$,
 then we receive a Hopfiled network (to get classical version we also need to convert binary values $\{0,1\}$ to $\{-1, 1\}$ and use real weights). 
So, the threshold model can be considered as a two-layer generalization of a Hopfield network that forms a bipartite graph.
It is an interesting research question to calculate the number of attractors in the presented models depending on the parameters. 
If instead of threshold nonlinearity we use sigmoid and Gibbs sampling to get binary activation, we would receive a well known Restricted Boltzmann Machine (RBM).

\section{Related works}
The problem of a binary representation of binary input with random binary weights was investigated before in works \cite{Hachemi2015,Hachemi2017}.
However, they used different activation function for the encoder: boolean operator OR, AND that is the same as a threshold model with $t_y=1$. This requires specific weight sparsity to get good results. The decoder is made by iterative optimization of augmented Lagrangian that gives better reconstruction compared to presented models here but requires more computation. Also, the models presented here are related to 1-bit compressed sampling, where the hidden layer is binary and the input is a vector with real numbers \cite{Jacques2013}.

Authors of the work \cite{Alemi2017} used a similar binary autoencoder setup, with fixed random weights of an encoder, but learned decoder weights. They analytically derived the capacity of the autoencoder ($\alpha_c = R / N_y $, where R - is the maximal number of patterns with zero error) using Gardner’s
 replica method for thermodynamic limit ($N_y \rightarrow \infty, N_x \rightarrow \infty$ but 
 the ratio $ N_y/N_x$ is limited).

The work \cite{Chaudhuri2019} deserves special attention. Authors connected error-correcting codes from information theory with the Hopfield network and used
the bipartite variation of the Hopfield network and expander graph to compute the hidden state. Such setup is shown to have exponential capacity of attractors ($2^{\alpha N_x}$ for some $\alpha \in [0, 1]$) and good error correction.

The work \cite{Vidybida2017b} brings particular interest. Much smaller network size (only 9 neurons) with signal propagation delay and full connectivity exhibits 
complex spiking dynamics.
The authors showed the emergence of limit cycle attractors higher than length 2 (as we have in this work) that can be a good starting point for using autoencoders for sequences. 
 
It is worth mentioning the work by the Numenta team \cite{Cui2016a}. Spatial pooler is an encoder of inputs with arbitrary sparsity to fixed sparse binary representation. It uses simplified competitive Hebbian learning (the kWTA activation function) and homeostatic mechanisms that tune the weights according to input statistics. It would be interesting to see how  biologically inspired mechanisms of learning influence mutual information.

Lastly, in the work \cite{Osaulenko2019} it was shown the improvement of association memory capacity by using the sigma-pi neuron model. The model was connected with the biological phenomenon of synaptic clustering when synapses from correlated inputs make connections on local dendritic branches. Current work provides
a deeper explanation of the previous results by analyzing the mutual information.

\section{Questions and discussion}
It turned out that the binary autoencoder model has links to many other fields. 
The terminology of the title is from the classical neural networks, but the work relates to sparse coding,
neuroscience, information theory, and data compression.

Here we used the uniform input distribution, so compression here is impossible.
But the next natural step is to investigate a non-uniform case for artificially generated data or some dataset.
In this case, the learning algorithm is needed, but the binary activation and weights make the problem quite complicated. 
Recently, it was shown \cite{Illing2019} the possibility to achieve a great performance with one hidden layer on a classification task with biologically plausible learning rules. 
But what is the goal of learning in autoencoders? Reduction of a training error for better compression? Or reduction of a test error for a better generalization? Or an ability to generate new examples that match the input distribution?
We have seen that the autoencoder forms the attractors. How learning is related to a search for the optimal number of attractors and their basins for the given data and model? 
How learning of invariants(like translation and scale for images) in the data might help with information preservation? 

Classical sparse coding assumes that the sparse input $\vec{x}$ 
can be reconstructed $\vec{x} = \matr{w}\vec{y}$ by a small number of elements from overcomplete dictionary ($N_y > N_x$).
Typically, the linear function and real values are considered with the goal to solve $P_0^\epsilon$ problem :
$ min(|\vec{y}|_0) s.t. ||\vec{x} - \matr{w}\vec{y}||_2 \leq \epsilon $ for some $\epsilon > 0$. 
For a classical case, to find the sparsest $\vec{y}$, two approaches are used: relaxation (for $L_1$ norm) and greedy search (for $L_0$ "norm") \cite{Elad2010,Elad2015}.     
In our case, the vectors are binary (note that $L_0$  is the same as $L_1$),
thus we need to use nonlinear reconstruction.
In this paper it is shown that for binary values it is possible to get sparse reconstruction $a_y \ll N_x$ for any input, 
not just sparse, given large enough dictionary. The major difference (and a drawback) is that knowing only the hidden state in not enough to get the unique input (it depends on the parameter $t_x$).
So, either we need to use a fixed learned $t_x$ for some dataset, or to encode the information about the $a_x$ somehow into the hidden state. 
  
The presented model is related to a simple biological network of a fruit fly, but how  is it justifiable? 
The real neuron emits spikes in continuous time, but here we assumed that time is discrete. 
It is possible to select some time window and artificially discretize the time, but still, it gives just an approximation.  It is hard to state perfect correspondence and most likely the computational experiments with spiking neural networks are needed. 

This work shows that the difference between sparse and dense representations can be explained through the memory-computation trade-off. 
In terms of information preservation, the sparse activation with large layers and simple models is the same as dense activation with smaller layers and more complex models.
The sparse activation observed not everywhere in the brain, there are regions with dense activity. Even in the one area of the neocortex, different layers can have different activation levels (for example L2/3 is sparse and L5/6 is dense \cite{Harris2012}).
In the fruit fly, the dynamic threshold that realized kWTA is set by a single inhibitory neuron that connects to excitatory neurons (our hidden layer).
However, in the neocortex, there are dozens of different inhibitory neuron types \cite{Markram2004}. 
The balanced inhibition \cite{Haider2006,Sprekeler2017} with the complex inhibitory and disinhibitory circuits \cite{Barron2017a} can complicate the 
commonly assumed weighted summation to a highly nonlinear model.
This could be the way to preserve the information with the dense activity.
Also, the architecture of a real neural network is governed not only by the memory-computation trade-off.  The balance between the model size (number of neurons) and the model complexity (the size of the neuron and the supporting machinery like the glial cells) is maintained by the physical limitations like the energy and material resources. 

For the human brain, the primary visual area V1 is 40 times larger than the input area from the thalamus (LGN)  \cite{VanEssen2005}.
It is much more than needed to preserve input information, and even more if we assume that inhibitory neuron diversity improves encoding. 
The typical proposal in the literature is that V1 does not process  only spatial pattern (like we studied in this paper)
but temporal patterns as well.
There is evidence of stable spatiotemporal patterns of neural activity in V1 in response to input sequences \cite{Gavornik2014,Carrillo-Reid2015}. 
This means that the probability distribution 
that the encoder should learn must include time $p(\vec{x_t},\vec{x_{t-1}, ...})$.
And hypothetically the learned spatiotemporal distribution would result in the ability to make correct predictions \cite{Keller2018} (that is essential for complex task solution and survival).
We can summarize all these thoughts in the question: how to maximize information across time?

The question how to form a sparse representation for arbitrary input is not fully answered. 
The full answer should consider the inputs from some spatiotemporal probability distribution.
A theory for a more general case should be developed.

\section{Appendix}
\subsection{Mathematical analysis}
To understand more deeply the encoding and decoding procedures of the threshold and kWTA models, we can analytically estimate the error of reconstruction. 

Let the overlap $z$ be some element from the vector $\vec{z} = \matr{w} \vec{x}$. The weights are random 
such as each row has exactly $a_w$ ones uniformly distributed.
In this case, the probability that overlap takes some specific value is:

\begin{equation}
p(z=k) = \frac{C^k_{a_x}C^{a_w-k}_{N_x-a_x}}{C^{a_w}_{N_x}}
\end{equation} 

\begin{figure}[t!]
  \begin{subfigure}[t]{0.49\textwidth}
        \includegraphics[height=2.5in]{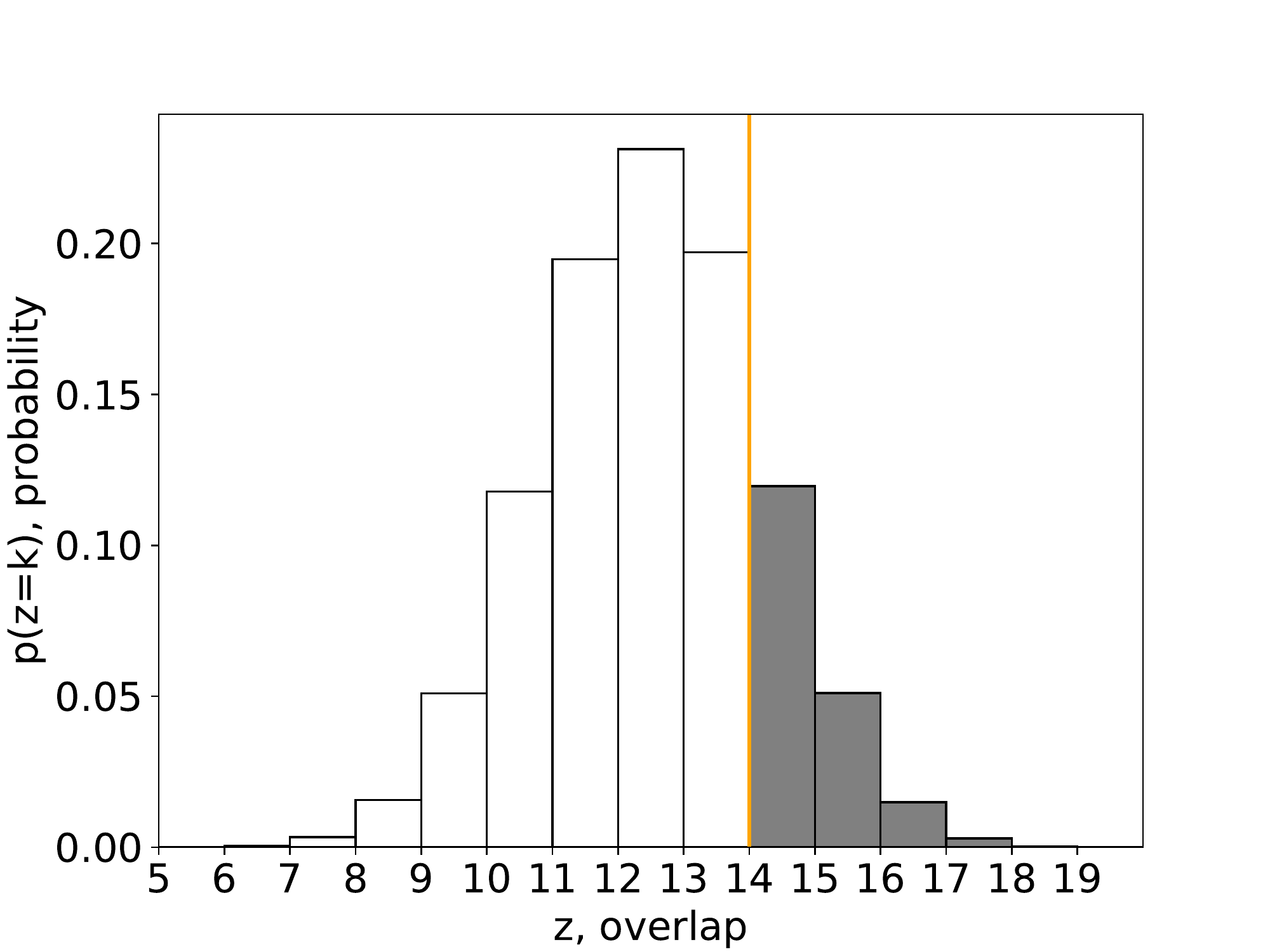}
      \caption{\label{fig:overlap_distribution} }  
  \end{subfigure}
  \qquad
  \begin{subfigure}[t]{0.49\textwidth}
      \includegraphics[height=2.5in]{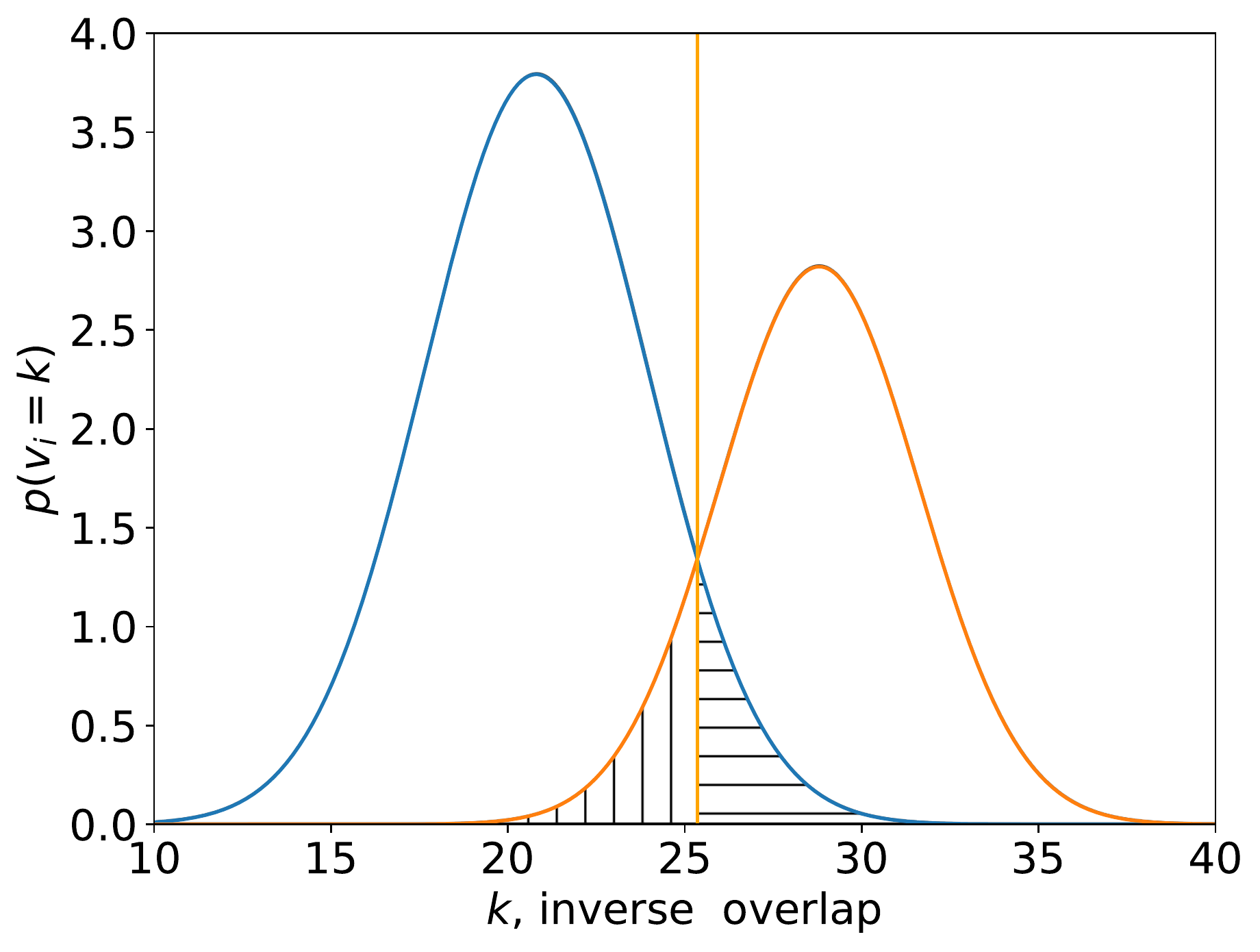}
      \caption{\label{fig:two_gaussians}}
  \end{subfigure}
     \caption{
          a) Distribution of overlap $z$,  for the $N_x=50, a_x = 20, N_y = 200, a_w = 30$. Vertical line shows the threshold $t_y=14$. 
          b)Illustration the distribution of overlap for the input layer ($v$).
           Left gaussian shows the distribution for cells $x_i = 0$, the right for $x_i = 1$.
           Vertical line shows the threshold $t_x$ that selects the true positive and the false positive cells.
          	}
\end{figure}

Where $C^k_{a_x}$ is the number we can chose $k$ ones from $a_x$. Also, we need to include the remaining $a_w-k$ 
ones among remaining zeros $N_x - a_x$ divided over the number of all possible weight vectors. 
The shape of the distribution is similar to a binomial and is presented on the fig. \ref{fig:overlap_distribution}.
The threshold $t_y$ defines the probability that the given cell is active $p(y_i = 1) = p(z \geq t_y)$ and is equal to the area of the shaded region.  

To find the reconstruction $\vec{x^r}$, let us define $\vec{v}=\matr{w}^T \vec{y}$ and repeat the procedure -- find its 
distribution and the area under the curve. Note, that alternatively, 
we can write $v_j = \sum_{i:y_i=1}{w_{ij}} $ as a sum of weight vectors from nonzero hidden cells (on average their number is $E(a_y) = N_yp(z \geq t_y)$). 
So, among all $N_y$ random vectors $\vec{w_i}$, we select those that have the overlap no less than the threshold $t_y$, sum them and apply the threshold $t_x$ to hopefully get the good input reconstruction. 

Next, let's analyze the weights that are used to reconstruct the input.  We still treat each vector $\vec{w_i}$  as a random, but we acquired information that it has $k$ overlapped ones with the input vector $\vec{x}$. It gives us the probabilities:

\begin{align}
& p^{+}(k) \equiv p(w_{ij}=1|x_j=1, z_i=k) = \frac{k}{a_x} \\
&p^{-}(k) \equiv p(w_{ij}=1|x_j=0, z_i=k) = \frac{a_w - k}{N_x -a_x}
\end{align}

So we can treat the $v_j$ as the sum of independent Bernoulli variables with the different parameters $p^{+}(k)$ (or $p^{-}(k)$), thus it is distributed according to Poisson binomial distribution. 
With a large number of weight vectors, this distribution is computationally intractable and we need to approximate it as a normal distribution. 
According to a central limit theorem:

\begin{align}
& p(v_j|x_j=1) \simeq N(\mu_{+}, \sigma_{+}^2) \\
& p(v_j|x_j=0) \simeq N(\mu_{-}, \sigma_{-}^2)
\end{align}

Where

\begin{align}
&\mu_{+} \approx \sum_{k=\theta_y}^{min(a_x,a_w)}{p^{+}(k)p(z=k)N_y} \\
&\sigma_{+}^2 \approx \sum_{k=\theta_y}^{min(a_x,a_w)}{p^{+}(k)(1-p^{+}(k))p(z=k)N_y}
\end{align}

And the same for $\mu_{-}, \sigma_{-}$ just plugging $p^{-}(k)$.  $p(z=k)N_y$ specifies the average number of cells that have the overlap equal to $k$.
For the strict equality for mean and variance, we need to sum over all possible combinations of overlap values, that is computationally hard (we would need to use combinations with repetition). 
These expressions are valid for the threshold model.
For the kWTA we need to adjust the sum, so that precisely $a_y$ additions remains (see details in the code \textit{fig6a.py} \cite{Github2020}).

The received two Gaussian are scaled by $N_x - a_x$ 
(left, $N(\mu_{-}, \sigma_{-}^2)$ ) and by $a_x$ (right) and are plotted on the fig. \ref{fig:two_gaussians}. 

The threshold $t_x$ (vertical line) for $\vec{v}$ defines the reconstruction vector $\vec{x^r}$.
The area with the vertical lines gives the average number of false negative neurons
$p(x^r_j=0|x_j=1)a_x$, cells that should be active but they do not. 
Area with the horizontal lines gives the number of false positive neurons $p(x^r_j=1|x_j=0) (N_x - a_x)$. 
Together they give the total reconstruction error.

So, the average Hamming distance:
\begin{equation}
E(d(\vec{x}, \vec{x^r})) = p(x^r_j=0|x_j=1)a_x + p(x^r_j=0|x_j=1)(N_x - a_x)
\end{equation}

As you can see, it decomposes on two parts, the false positive and false negative.
Indeed, the original expression can be rewritten as 
\begin{equation}
\sum_i{|x_i - x^r_i|}  = \vec{x}^T(\mathbf{1} - \vec{x^r} ) + (\mathbf{1} - \vec{x})^T \vec{x^r}
\end{equation}

\begin{figure}[t!]
\begin{subfigure}[t]{0.49\textwidth}
        \includegraphics[height=2.5in]{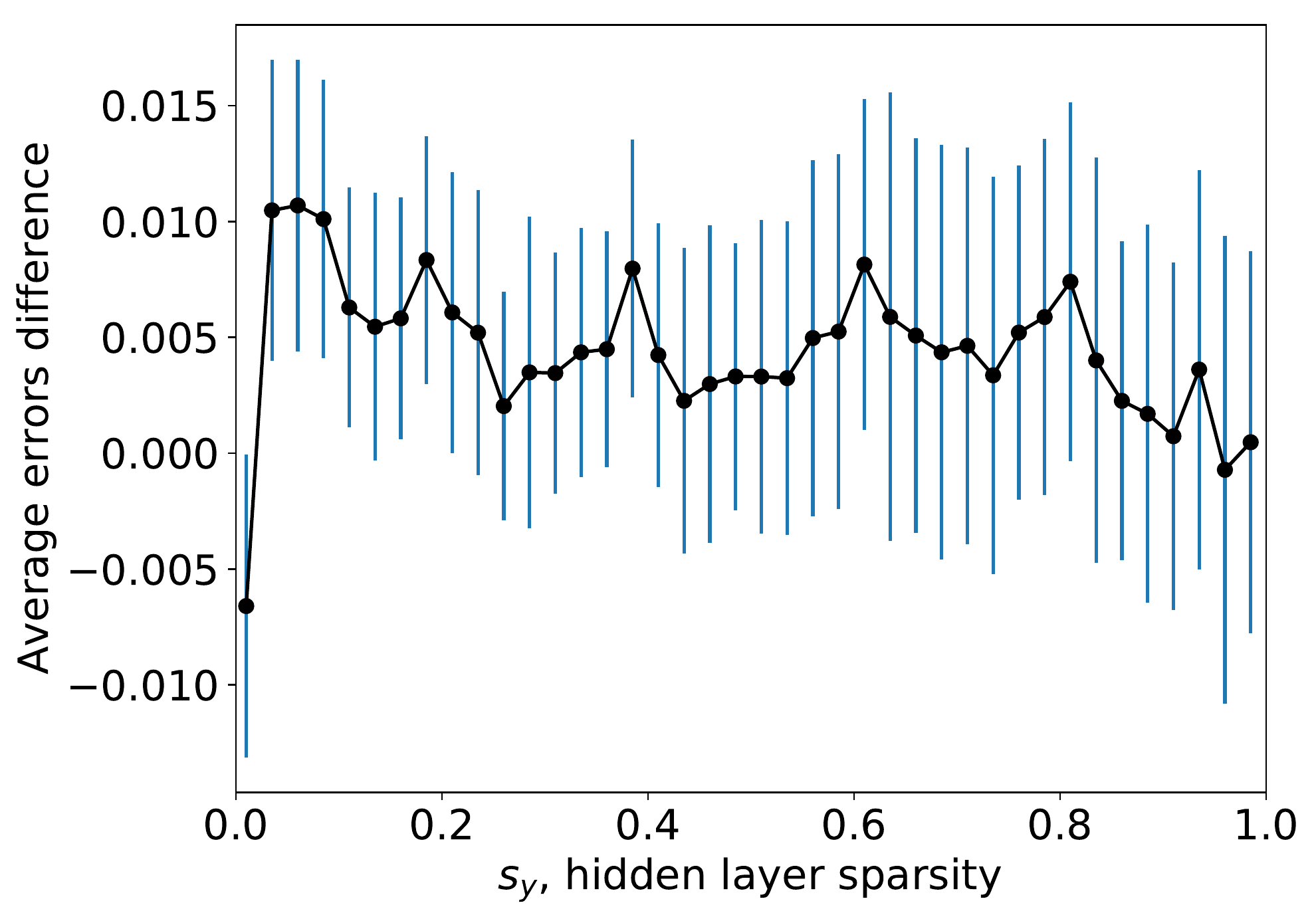}
      \caption{\label{fig:error_vs_sparsity} }  
  \end{subfigure}
  \qquad
  \begin{subfigure}[t]{0.49\textwidth}
      \includegraphics[height=2.5in]{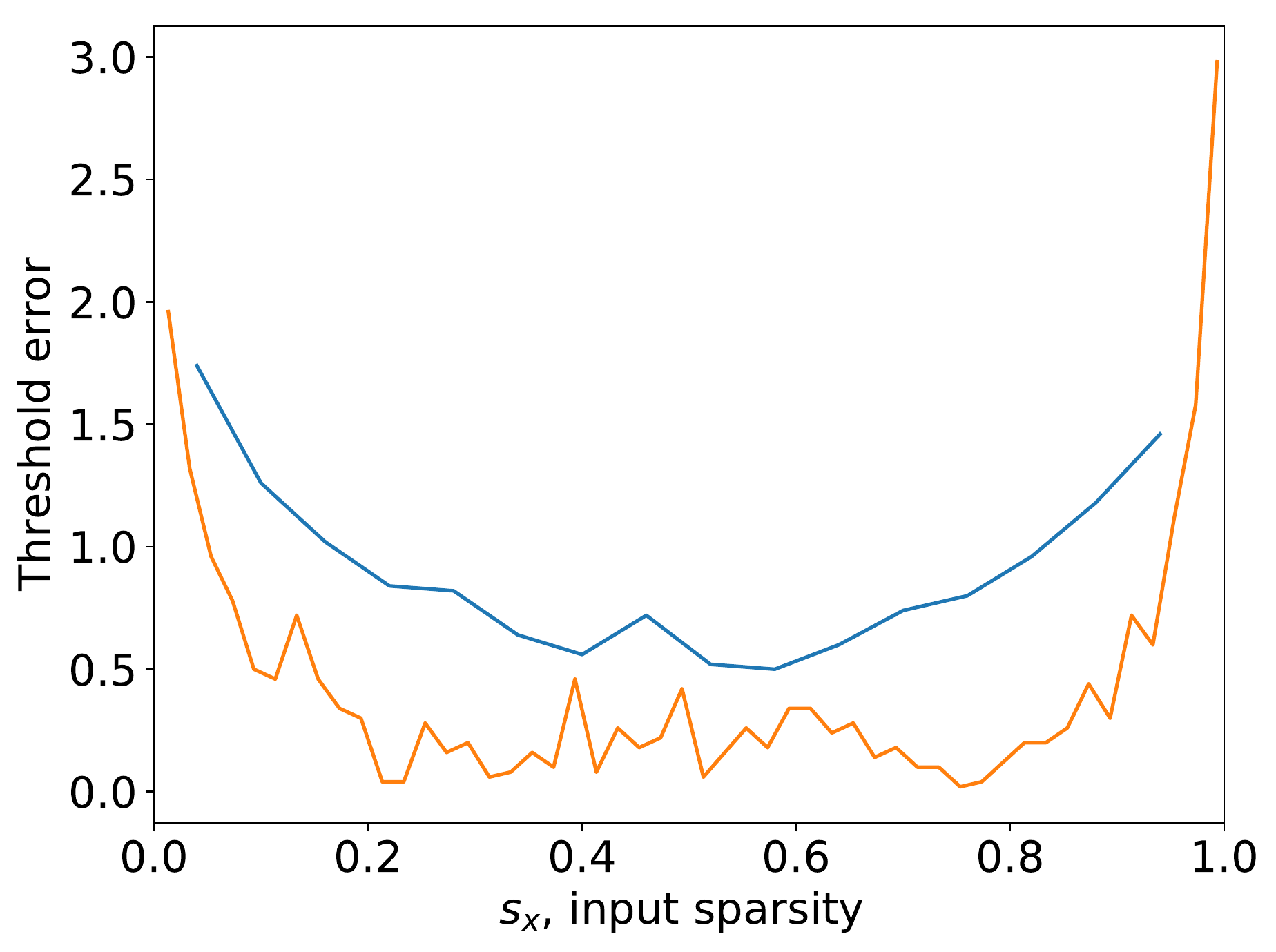}
      \caption{\label{fig:thresh_error}}
  \end{subfigure}
     \caption{
          a)  Difference of experimental average error and the theoretical estimation depending on hidden layer sparsity.
           Vertical lines show the standard deviation. The difference is larger for a small number of active hidden layer cells and tends to zero for larger.
          b) Difference of experimental average optimal threshold and its approximation depending on input layer sparsity. 
          Upper curve is for the network with $N_x=50, N_y=200, a_w=30$, lower for $N_x=100, N_y=300, a_w=60$.
           The error difference is the lowest for $s_x$ close to 0.5.  
          	}
\end{figure}

The presented analysis shows, 
that for random binary weights and binary activations it is possible to get the mean error analytically without the need to run autoencoder. 
To see how accurate the theoretical result with the applied approximations, we need to compare it with computational experiments.
The fig.  \ref{fig:error_vs_sparsity} shows the difference between theoretical and computational average error depending on the hidden layer sparsity.
As we can see the difference is quite small, less than 1\% and gets smaller with the increase of $s_y$.
The more accurate results for more active hidden layer neurons is due to the more accurate normal approximation of Poison binomial distribution.
The little spikes in the error difference correspond to values when the threshold $t_y$ changes.

\subsubsection{The thresholds}
Can the presented analysis give some intuition about how to select optimal thresholds?
Our goal is to find the parameters that reduce the overlap of two Gaussians( fig. \ref{fig:two_gaussians}). 

The threshold $t_y$ is in the range $[1, \min(a_x, a_w)]$ and it changes the distance between two Gaussians and their variances.
The first approximation of the optimal threshold can be calculated from the maximization of the difference of the means. After calculation we obtain:

\begin{equation}
t_y^1 = \argmaxG_{t_y} (\mu_{+} - \mu_{-}) = \frac{a_xa_w}{N_x} + 1
\end{equation}

The fig\ref{fig:thresh_error} shows the error of a threshold estimation $|t_y^1 - t_y^{c}|$ where $t_y^{c}$ 
is the average optimal threshold from computational experiments. Two curves correspond to two different parameter sets: upper ($N_x=50, N_y=200, a_w=30$), 
lower(($N_x=100, N_y=300, a_w=60$)). 
The first approximation based just on means of Gaussians gives not bad results, 
especially for larger networks and dense input vectors. For better estimation, we need to take into account variances and more carefully use a normal approximation.
Also, note that this analysis gives an average estimation of the optimal threshold, but the optimal value for each individual weight setup may be different. 

Two Gaussians on fig. \ref{fig:two_gaussians} are scaled to graphically show that the optimal 
threshold $t_x$ should be selected to minimize the area under two opposing gaussian tales.
The optimal threshold for the input layer can be found from:

\begin{equation}
t_x = \argminG_{t_x}(\Phi(\frac{t_x - \mu_{+}}{\sigma_{+}})a_x - \Phi(\frac{t_x - \mu_{-}}{\sigma_{-}})(N_x - a_x))
\end{equation}

Where $\Phi(\cdot)$ is a cumulative distribution function of a normal distribution.
Note that this value is not an integer.
Perhaps we would need for more rigorous analysis to discretize the Gaussians and apply 1/2 correction, or any other corrections. 
However, with the Gaussians, the error estimation is still pretty good (see fig.\ref{fig:error_vs_sparsity}).

\subsection{Model variations}
\label{model_variations}
The presented models may have many different variations. 
For example, it is interesting to test what will be if activation and weigh values are $\{-1, 1\}$, or even ternary  $\{-1, 0, 1 \}$.
Or consider random input $x_i \in Ber(s_x)$.

Throughout the paper, we used a special distribution of weights where the row has exactly $a_w$ ones. 
The results show that it gives better results compared to Bernoulli distribution $w_{ij} \in Ber(s_w)$
or even the real values from uniform or standard normal distributions (see the code \textit{weigths\_comparison.py}). 
The threshold or kWTA function selects some number of ordered weights to reconstruct the input.
The criterion for ordering is a dot product $\matr{w}\vec{x}$ with the input. 
For this case, 
the reconstruction is more accurate when the selected weights 
are the most similar to inputs.
In the case of fixed number of nonzero weights in a row, the dot product gives the same ordering as the cosine similarity ($cos(\vec{w_i}, \vec{x}) = \frac{\vec{w_i} \cdot \vec{x}}{||\vec{w_i}||_2 ||\vec{x}||_2 } $) since all rows have the same value of L2 norm $||\vec{w_i}||_2$. 
For other cases (like the Bernoulli), the ordering is different (because the norm is different) and the chosen weights to reconstruct the input are not the most similar. 
But if we change the activation function to $f(cos(\vec{w_i}, \vec{x}))$, the weight from different distribution gives the same result.
Also, another reason to chose the fixed number of nonzero weights, is that it gives a much simpler mathematical analysis, where we do not need to calculate many averages.

Consider another variation, when every neuron has an individual threshold (bias) like in the classical artificial neural networks (ANN). 
Will individual threshold improve the reconstruction since the model has more parameters? No. 
The classical ANN have a different approach: given the train data the model learns parameters, fix them and evaluate performance for the test data.
In this paper the shared threshold is not fixed and depends on every input. 
The optimal individual threshold is the averaged values of the shared threshold for some neuron.
However, for some input the fixed individual threshold is different from the threshold that minimized the reconstruction error. Thus, the dynamic shared threshold gives better results.

\subsection{More intuition of reconstruction}
\label{intuition_reconstruction}
Suppose we need to find a reconstruction of any binary input vectors of size $N_x$.
We could construct the dictionary where each row contains 
only one nonzero element and all rows are different. There are $C^1_{N_x} = N_x$ possible rows. Such dictionary of size $N_x \times N_x$ will give the reconstruction with at most $a_y = a_x$ nonzero elements in $\vec{y}$ vector (input just reconstructed element by element). 
We could extend the dictionary with the rows with all pairwise combinations of nonzeros ($C^2_{N_x}$). In this case we need at most $a_y = a_x / 2$ nonzero elements in $\vec{y}$. We could extend the dictionary further with more combinations and reduce the sparsity of reconstruction even more up to $a_y = 1$.
But it requires a huge dictionary. Much more interesting case is when we have the limited dictionary (for example $N_y = R$ ) and nonuniform input distribution. 
We can start with the full dictionary of size $N_y = 2^{N_x}$ and ask ourselves: what elements we need to throw away to make the size equal to $R$ but preserve that lowest error and sparsest reconstruction? 
Intuitively, we need to remain those elements that correspond to the most probable inputs. For example, if some pairwise correlation ($x_5x_{12}$) is absent in the input distribution, we do not need the corresponding row where 5-th and 12-th elements are ones. 
Maybe, such a process of dictionary reduction somehow corresponds to what is known as neural and synaptic pruning at early stages of neural development when many neurons and synapses are eliminated.

\subsection{mAP calculation }
\label{similarity_map}
To calculate the mean Average Precision (mAP) for the presented models the following procedure was performed: 
\begin{enumerate}
\item 1000 random binary vectors of size $N_x$ with fixed number of ones $a_x$ was generated;
\item for some selected vector $\vec{x_s}$ the 20 closest in L1 distance vectors were found;
\item for the encoded vector $\vec{y_s} = f(\vec{x_s})$   20 closest in the hidden space vectors were found;
\item  the average precision was calculated as follows:
 let $A$ be an ordered set of indices of vectors similar to $\vec{x_s}$ in the input space, $B=\{b_i|i=1:20\}$ is an ordered set of indices of similar vectors to $\vec{y_s}$ in the hidden space, $I_A(x)= \{1, \text{if } x \in A; 0, \text{othervise} \}$ is  an indicator function, then the
 average precision is: 
 \begin{equation}
 ap = \sum_{i=1}^{20}{\frac{I_A(b_i)}{i}}
 \end{equation}
  It counts the number of vectors that belong  to closest set both in the input and the hidden space and is adjusted by the ordering of the positive retrievals.
\item the selection of different vector from $\vec{x_s}$ was repeated 100 times and for each  the average precision was received. 
\item  mean averaged precision was obtained by averaging 100 random input vector selections.
\end{enumerate}

To get the dependence of mAP on the hidden layer sparsity, for each value $s_y$ 
the mAP was calculated 10 times, each time with different random weights,  and averaged.

\subsection{Mutual information derivation}
\label{mutual_info}
Let $X$ represent the input state and $Y$ the hidden state.  For simplicity the following notation is used: $p(Y=y_i)=p(y_i)$ and the same for $X$.
\begin{equation}
\begin{aligned}
\begin{split}
&I(X,Y) = H(Y) \euqall{1} \sum_{i=1}^Z{p(y_i)log_2(p(y_i))} \\
& \euqall{2} -\sum_{i=1}^Z{\sum_{j=1}^{2^{N_x}}{p(y_i| x_j)p(x_j)}log_2(\sum_{j=1}^{2^{N_x}}{p(y_i| x_j)p(x_j)})} \\
& \euqall{3} -\sum_{i=1}^Z{\sum_{j \in \Omega_i} p(x_j) log_2(\sum_{j \in \Omega_i} p(x_j)}) \\ 
& \euqall{4} -\sum_{i=1}^Z{\frac{|\Omega_i|}{2^{N_x}}log_2(\frac{|\Omega_i|}{2^{N_x}})} \\ 
& \euqall{5} N_x - \frac{1}{2^{N_x}}\sum_{i=1}^Z{|\Omega_i| log_2(|\Omega_i|)} \\ 
& \less{6} N_x - log_2(\Omega^*) \\ 
\end{split}
\end{aligned}
\end{equation}

At (1) $Z = 2^{N_x}/E(|\Omega_i|)$ -- all realised vectors, 
and $\Omega_i = \{\vec{x_j}| \vec{y_i} = f(\vec{x_j}) \}$ -- a set of input vectors that lead to one hidden vector $\vec{y_i}$ ,
  $|\Omega_i|$ - number of elements in the set
 and $\Omega^* = E(|\Omega_i|)$ the mean number.  

At (2) the law of total probability .

At (3) the hidden state is the deterministic function of input $p(y_i| x_j) = 1$  if $x_j \in \Omega_i$ and 0 otherwise  (see fig.\ref{fig:encoder}). 

At (4) the $X$ is assumed to be uniform ($p(x_i) = 1/2^{N_x}$). 

At (5) use that $\sum_i{|\Omega_i|} = 2^{N_x}$ and rearrange the logarithm. Also, note that for our case $H(X) = N_x$.

 At (6) apply log sum inequality :

\begin{equation}
\begin{aligned}
\begin{split}
&\sum_{i=1}^Z{|\Omega_i| log_2(|\Omega_i|)} \euqall{1} \sum_{i=1}^Z{|\Omega_i| log_2(\frac{|\Omega_i|}{b_i})}
\geq (\sum_{i=1}^Z{|\Omega_i|}) log_2(\frac{\sum_i{|\Omega_i|}}{Z}) 
&  \euqall{2} 2^{N_x} log_2(\Omega^*)
\end{split}
\end{aligned}
\end{equation}

At (1) divide by the ones $\vec{b} = \vec{1}$ to directly use log sum inequality that states $\sum_i{a_ilog(\frac{a_i}{bi})} \geq alog(\frac{a}{b})$ where $a_i, b_i$ non-negative numbers and $a = \sum_i{a_i}, b = \sum_i{b_i}$.

At (2) use $Z = \frac{2^{N_x}}{\Omega^*}= \frac{\sum_i{|\Omega_i|}}{\Omega^*}$

For general case, to increase mutual information, we would need to increase $Z$ and to make $\sum_{j \in \Omega_i} p(x_j)\rightarrow \Psi$  for some $\Psi$ for all $\Omega_i$.

\bibliographystyle{unsrt}  
\bibliography{reference}  %%% Remove comment to use the external .bib file (using bibtex).
%%% and comment out the ``thebibliography'' section.

%%% Comment out this section when you \bibliography{references} is enabled.
%\begin{thebibliography}{1}
%
%\bibitem{kour2014real}
%George Kour and Raid Saabne.
%\newblock Real-time segmentation of on-line handwritten arabic script.
%\newblock In {\em Frontiers in Handwriting Recognition (ICFHR), 2014 14th
%  International Conference on}, pages 417--422. IEEE, 2014.
%
%\bibitem{kour2014fast}
%George Kour and Raid Saabne.
%\newblock Fast classification of handwritten on-line arabic characters.
%\newblock In {\em Soft Computing and Pattern Recognition (SoCPaR), 2014 6th
%  International Conference of}, pages 312--318. IEEE, 2014.
%
%\bibitem{hadash2018estimate}
%Guy Hadash, Einat Kermany, Boaz Carmeli, Ofer Lavi, George Kour, and Alon
%  Jacovi.
%\newblock Estimate and replace: A novel approach to integrating deep neural
%  networks with existing applications.
%\newblock {\em arXiv preprint arXiv:1804.09028}, 2018.
%
%\end{thebibliography}

\end{document}